%% file: smogu_arxiv.tex
\documentclass{article}

\usepackage{arxiv}
\usepackage[utf8]{inputenc}
\usepackage[T1]{fontenc}
\usepackage{hyperref}
\usepackage{url}
\usepackage{booktabs}
\usepackage{graphicx}
\usepackage{natbib}
\usepackage{microtype}
\usepackage{float}
\usepackage{tikz}
\usepackage{subcaption}
\usepackage{xcolor}
\usetikzlibrary{shapes.geometric,arrows.meta,positioning}
\graphicspath{{images/}{plots/}}

\hypersetup{
  hidelinks,
  pdftitle={Cross-Entropy Guided Routing in Mixture-of-Experts Large Language Models},
  pdfauthor={Yury Nahshan, Nati Daniel, Jacob Goldberger, Yoli Shavit},
  pdfkeywords={Mixture-of-Experts, sparse routing, token-error supervision, language models}
}

\input{math_commands.tex}

\title{Cross-Entropy Guided Routing in Mixture-of-Experts Large Language Models}
\author{
Yury Nahshan$^{1,2}$ \quad
Nati Daniel$^{2}$ \quad
Jacob Goldberger$^{1}$ \quad
Yoli Shavit$^{1}$\\[0.5em]
$^{1}$Bar-Ilan University, Ramat-Gan, Israel\\
$^{2}$NVIDIA, Israel
}
\date{}
\renewcommand{\shorttitle}{Cross-Entropy Guided Routing in Mixture-of-Experts Large Language Models}

\begin{document}
\maketitle

\begin{abstract}
Sparse mixture-of-experts (MoE) large language models scale model capacity by routing
each token to a small subset of experts.
Their routers are regularized with load balancing terms and learn affinity scores
through the language-model objective. However, these objectives do not provide
direct alignment between routing affinities and token-level error.
We introduce token-error supervision
for sparse routing in two forms. The first form predicts an error score per
expert. The affinity-weighted aggregate of these scores is aligned to the
next-token cross-entropy loss, while the individual scores attenuate affinity
before top-$K$ selection. The second directly aligns the router's affinities to
the model's objective without requiring an additional head or inference-time
modification.
Both formulations use the Itakura--Saito divergence or an exponential negative
log-likelihood for aligning affinities and token errors. Across two sparse MoE
backbones and four multiple-choice question-answering benchmarks, we evaluate
both supervision mechanisms. On Granite, our method improves accuracy by
approximately 2.3 percentage points on average over a parameter-matched routing
baseline. With stronger supervision, the gain on ARC-Challenge reaches 2.94
points. Both mechanisms preserve the native sparse execution budget and
aggregation policy. Our code is available in the supplementary materials.
\end{abstract}

\input{chapters/introduction}
\input{chapters/related_work}
\input{chapters/method}
\input{chapters/results}
\input{chapters/conclusion}


\bibliography{smogu}
\bibliographystyle{iclr2027_conference}

\clearpage
\appendix
\input{chapters/appendix}

\end{document}

%% file: math_commands.tex
\usepackage{amsmath,amsfonts,bm}

\def\eqref#1{equation~\ref{#1}}

\def\1{\bm{1}}

\DeclareMathAlphabet{\mathsfit}{\encodingdefault}{\sfdefault}{m}{sl}
\SetMathAlphabet{\mathsfit}{bold}{\encodingdefault}{\sfdefault}{bx}{n}



%% file: chapters/introduction.tex
\section{Introduction}\label{sec:introduction}

Sparse mixture-of-experts (MoE) models expand model capacity without
proportionally increasing computation by routing each token to only $K$ of $N$
available experts \citep{shazeer2017outrageously,lepikhin2021gshard,
fedus2022switch}. This conditional computation has become central to scaling
modern language models \citep{jiang2024mixtral,dai2024deepseekmoe}. Its
effectiveness, however, depends on routing: for every token, the router
determines which experts are executed and how their outputs
contribute to the token representation.
Standard sparse routers make this decision through affinity scores learned
indirectly from the language-model objective and routing regularizers such as
load balancing \citep{fedus2022switch}. These scores provide relative
preferences among experts, but they are not explicitly supervised against the
model's token-level loss. Native MoE routing therefore lacks an explicit signal
indicating whether the selected computation is likely to produce a high- or
low-error prediction.

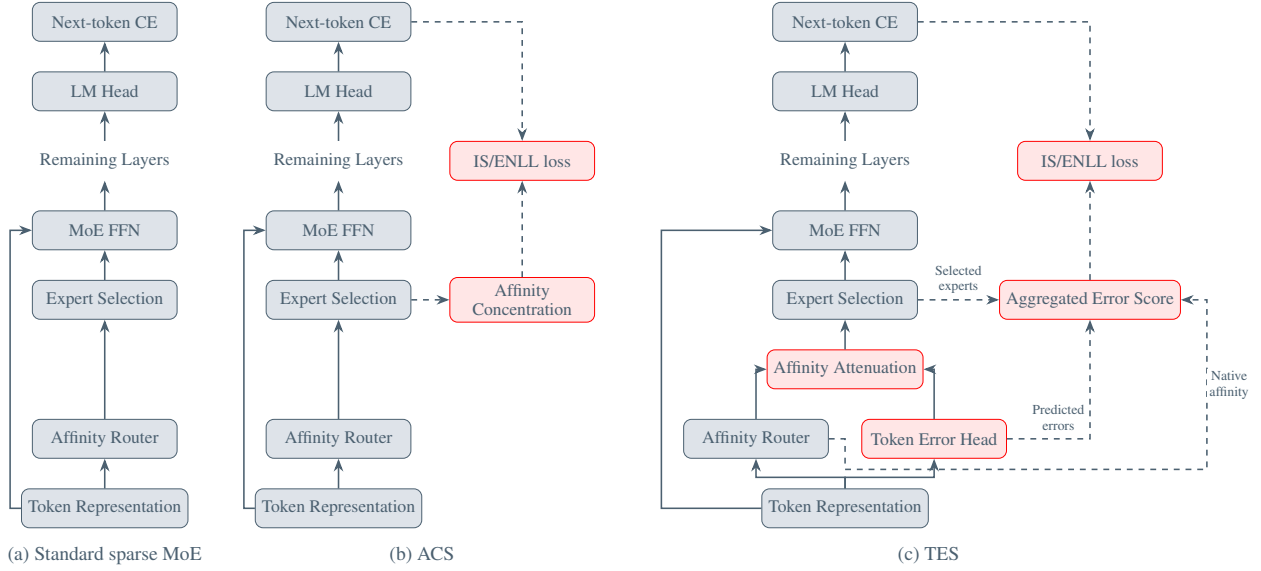
\begin{figure}[t]
    \centering
    \input{images/figure1_tikz}
    \caption{Overview of sparse MoE routing and the proposed supervision
    mechanisms: (a) the standard language-model pipeline; (b) affinity-concentration
    supervision (ACS); and (c) token-error supervision (TES) with affinity
    attenuation. MoE FFN denotes the routed expert computation and aggregation;
    routing supplies expert selection and aggregation weights. ACS computes
    concentration from normalized selected affinities. TES aggregates predicted
    errors using native affinities normalized over the executed experts.
    Dashed arrows denote auxiliary readout inputs,
    not stopped gradients.}
    \label{fig:method_overview}
\end{figure}

The realized next-token cross-entropy provides a direct token-level measure of
prediction error for the routed computation. We study two ways of using this
signal. The first, token-error supervision (TES), trains an error prediction
head and aggregates the predicted errors based on affinity scores.
Figure~\ref{fig:method_overview} compares standard sparse MoE routing with
the two supervision mechanisms. The second,
affinity-concentration supervision (ACS), uses the realized token loss to
supervise the squared probabilities of the normalized selected-expert
affinities, allowing training to sharpen or flatten the native routing
distribution without an additional head or inference-time transformation. We
supervise TES and ACS with the
Itakura--Saito (IS) divergence
\citep{itakura1968analysis,fevotte2009nonnegative}. We further explore the exponential negative
log-likelihood (ENLL) \citep{casella2002statistical} as a simplified supervision scheme.

Our theoretical analysis shows that, for TES, both objectives train the
aggregate prediction toward the same expected token-error target, but affect language-model optimization differently.
When predicted and observed token error match, IS leaves the original
cross-entropy (CE) update unchanged,
whereas ENLL gives relatively greater gradient weight to lower-loss tokens. In
TES, we subtract a scaled log-error score from native affinity before top-$K$
selection, reducing the relative preference of experts with larger predicted
error. In ACS, the same objectives instead sharpen or flatten native affinity
according to the token loss. Section~\ref{sec:method} develops the two mechanisms,
while Appendix~\ref{sec:appendix} analyzes their optima and gradients.

We evaluate TES and ACS on Granite 3.1 and OLMoE-1B across four
multiple-choice question-answering (MCQA) benchmarks. On Granite, ACS-IS
improves average accuracy by 2.1 percentage points over CE fine-tuning with
frozen native affinity. TES-IS improves over a parameter-matched Dual Affinity
baseline by 2.3 points on Granite and 0.5 points on OLMoE, averaged across all
four benchmarks. With stronger supervision, its improvement on Granite
ARC-Challenge reaches 2.94 points. These results show that directly supervising predicted token error or
affinity concentration against the realized token loss can improve sparse routing,
with the preferred mechanism depending on the backbone and task. Both methods
preserve the native top-$K$ execution budget and architecture-specific
aggregation policy.

In summary, our contributions are as follows:
\begin{itemize}
    \item
    We introduce two complementary ways to supervise sparse routing with
    realized next-token cross-entropy. A lightweight expert-indexed error head
    predicts token error through its native-affinity-weighted aggregate, while a headless alternative directly aligns native router outputs. We
    show that direct supervision enables token-error prediction and aligns affinity outputs with token error, without changing the sparse expert-execution budget.

    \item We further propose a
    method for integrating expert-indexed error scores into routing by
    subtracting a scaled log-error term from native affinity before top-$K$
    selection. The resulting route uses token-error information to rank experts
    while preserving the native execution budget and architecture-specific
    aggregation policy.

    \item We evaluate both supervision mechanisms against standard CE
    fine-tuning and a parameter-matched routing-head control across two sparse
    MoE backbones and four MCQA benchmarks, demonstrating accuracy improvements
    over the corresponding baselines.
\end{itemize}

%% file: images/figure1_tikz.tex
\definecolor{figuregray}{HTML}{DEE3E9}
\definecolor{figureink}{HTML}{4B6070}
\definecolor{figurered}{HTML}{FFE6E6}
\resizebox{\linewidth}{!}{%
\begin{tikzpicture}[
color=figureink,
box/.style={draw, rectangle, rounded corners, minimum width=2.6cm, minimum height=0.7cm, align=center, fill=figuregray, font=\small},
highlight/.style={draw=red, rectangle, rounded corners, minimum width=2.6cm, minimum height=0.7cm, align=center, fill=figurered, font=\small},
plain/.style={minimum width=2.6cm,minimum height=0.7cm,align=center,font=\small},
arrow/.style={-Stealth, thick},
dashedarrow/.style={-Stealth, thick, dashed},
note/.style={font=\scriptsize,fill=white,inner sep=1pt,align=center}]
\node[box] (a-token) at (0,0) {Token Representation};
\node[box] (a-sel) at (0,3.75) {Expert Selection};
\node[box] (a-ffn) at (0,5) {MoE FFN};
\node[plain] (a-rest) at (0,6.25) {Remaining Layers};
\node[box] (a-lm) at (0,7.5) {LM Head};
\node[box] (a-ce) at (0,8.75) {Next-token CE};
\draw[arrow] (a-sel)--(a-ffn);
\draw[arrow] (a-ffn)--(a-rest);
\draw[arrow] (a-rest)--(a-lm);
\draw[arrow] (a-lm)--(a-ce);
\draw[arrow] (a-token.west)--(-1.7,0)--(-1.7,5)--(a-ffn.west);
\node[box] (a-aff) at (0,1.25) {Affinity Router};
\draw[arrow] (a-token)--(a-aff);
\draw[arrow] (a-aff)--(a-sel);
\node[box] (b-token) at (4.2,0) {Token Representation};
\node[box] (b-sel) at (4.2,3.75) {Expert Selection};
\node[box] (b-ffn) at (4.2,5) {MoE FFN};
\node[plain] (b-rest) at (4.2,6.25) {Remaining Layers};
\node[box] (b-lm) at (4.2,7.5) {LM Head};
\node[box] (b-ce) at (4.2,8.75) {Next-token CE};
\draw[arrow] (b-sel)--(b-ffn);
\draw[arrow] (b-ffn)--(b-rest);
\draw[arrow] (b-rest)--(b-lm);
\draw[arrow] (b-lm)--(b-ce);
\draw[arrow] (b-token.west)--(2.5,0)--(2.5,5)--(b-ffn.west);
\node[box] (b-aff) at (4.2,1.25) {Affinity Router};
\draw[arrow] (b-token)--(b-aff);
\draw[arrow] (b-aff)--(b-sel);
\node[highlight] (b-conc) at (7.5,3.75) {Affinity\\Concentration};
\node[highlight] (b-loss) at (7.5,6.25) {IS/ENLL loss};
\draw[dashedarrow] (b-sel.east)--(b-conc.west);
\draw[dashedarrow] (b-conc)--(b-loss);
\draw[dashedarrow] (b-ce.east)-|(b-loss.north);
\begin{scope}[xshift=1.1cm]
\node[box] (c-token) at (12.2,0) {Token Representation};
\node[box] (c-sel) at (12.2,3.75) {Expert Selection};
\node[box] (c-ffn) at (12.2,5) {MoE FFN};
\node[plain] (c-rest) at (12.2,6.25) {Remaining Layers};
\node[box] (c-lm) at (12.2,7.5) {LM Head};
\node[box] (c-ce) at (12.2,8.75) {Next-token CE};
\draw[arrow] (c-sel)--(c-ffn);
\draw[arrow] (c-ffn)--(c-rest);
\draw[arrow] (c-rest)--(c-lm);
\draw[arrow] (c-lm)--(c-ce);
\draw[arrow] (c-token.west)--(8.9,0)--(8.9,5)--(c-ffn.west);

\node[box] (c-aff) at (10.6,1.25) {Affinity Router};
\node[highlight] (c-error) at (13.8,1.25) {Token Error Head};
\node[highlight] (c-atten) at (12.2,2.5) {Affinity Attenuation};
\node[highlight] (c-agg) at (16.6,3.75) {Aggregated Error Score};
\node[highlight] (c-loss) at (16.6,6.25) {IS/ENLL loss};
\draw[arrow] (c-token.north)--++(0,0.2)-|(c-aff.south);
\draw[arrow] (c-token.north)--++(0,0.2)-|(c-error.south);
\draw[arrow] (c-aff.north)|-(c-atten.west);
\draw[arrow] (c-error.north)|-(c-atten.east);
\draw[arrow] (c-atten)--(c-sel);
\draw[dashedarrow] (c-sel.east)--node[note,above=2pt]{Selected\\experts}(c-agg.west);
\draw[dashedarrow] (c-error.east)-|node[note,near start,above=3pt,xshift={8pt-1mm}]{Predicted\\errors}(c-agg.south);
\draw[dashedarrow] (c-aff.east)--(12.2,1.25)--(12.2,0.7)--(18.7,0.7)
--node[note,right]{Native\\affinity}(18.7,3.75)--(c-agg.east);
\draw[dashedarrow] (c-agg)--(c-loss);
\draw[dashedarrow] (c-ce.east)-|(c-loss.north);
\node at (13.7,-0.85) {(c) TES};
\end{scope}
\node at (0,-0.85) {(a) Standard sparse MoE};
\node at (5.7,-0.85) {(b) ACS};
\end{tikzpicture}%
}

%% file: chapters/related_work.tex
\section{Related Work}\label{sec:related_work}

\paragraph{Sparse MoE routing and explicit supervision.}
Sparse MoE models use learned affinity to activate only the top-$K$ experts,
increasing capacity without proportional per-token computation
\citep{shazeer2017outrageously,lepikhin2021gshard,fedus2022switch}. Later work
changes how this sparse allocation is formed: Expert Choice lets experts select
tokens \citep{zhou2022expertchoice}, while ReMoE replaces discontinuous top-$K$
selection with differentiable ReLU routing \citep{wang2025remoe}.

A closer line of work explicitly guides router behavior. Expert-router coupling (ERC) binds router
embeddings with expert capabilities through expert-specific proxy tokens
\citep{lv2026erc}, and Expert Divergence uses domain labels to encourage
functional specialization \citep{li2026expertdivergence}. Counterfactual
analysis further shows that native routing can miss equal-compute alternatives
with lower next-token loss \citep{yoon2026misrouted}. Our method directly
targets this gap through two complementary mechanisms. TES supervises an
affinity-weighted prediction of routed token loss and uses the resulting
expert-indexed error scores to refine selection without changing the top-$K$
execution budget. ACS uses the same token-loss objectives to shape affinity without adding a prediction head.

\paragraph{Token difficulty and loss prediction.}
Learned loss prediction provides a general mechanism for estimating which
inputs a model is likely to find difficult \citep{yoo2019learning}. At the query
level, Hybrid LLM uses predicted difficulty to route requests between models of
different capacities \citep{ding2024hybrid}. Within MoE models,
\citet{huang2024harder} infer difficulty from router confidence and activate
more experts for difficult inputs; DynaMoE derives token-difficulty labels from
agreement between nested experts and the full-width MLP
\citep{nishu2025dynamoe}; and Ada-K learns a token-dependent expert budget
through reinforcement learning \citep{zhao2025adak}. These methods use
difficulty to allocate computation, typically by changing the capacity or
number of experts assigned to each token.
Our method instead uses realized next-token loss to supervise sparse language-model
routing. 

\paragraph{Uncertainty-aware routing.}
In sparse language models, recent work represents routing uncertainty through
different probabilistic signals. Variational Mixture-of-Experts Routing
(VMoER)~\citep{li2026vmoer} represents routing logits using input-dependent
probability distributions and performs variational inference over the
resulting routing decisions. Uncertainty-Aware Routing
(UAR)~\citep{chen2026uar} uses router entropy to adapt both expert capacity and
routing regularization.
Related probabilistic routers include Grassmannian MoE
\citep{shihab2026grmoe}, which controls routing through Bingham
subspace geometry, while VI-MoLE \citep{saliencro2026vimole} predicts
counterfactual residual risk to allocate a variable budget among LoRA experts.
Our method uses deterministic, fixed-budget top-$K$ routing without sampling
router distributions and introduces direct supervision from the routed
model's next-token loss. Fixed-budget execution is also retained by VMoER's
logit-space inference; our distinction is the directly supervised error signal.

In dense MoE models for time-series regression, MoGU \citep{aviv2025mogu}
provides the closest conceptual precedent for using an expert-specific
predictive-error signal to control mixture weights. It
models each regression expert as a Gaussian predictor, trains its variance
through Gaussian negative log-likelihood, and aggregates expert predictions
using normalized inverse variance. Our formulation differs in both target and
routing semantics: the error head does not define a Gaussian expert likelihood
or estimate calibrated predictive variance. Instead, native affinity aggregates
the executed experts' positive scores into a token-error prediction supervised
against next-token cross-entropy; the scaled log-error values subsequently
reduce the relative affinity of experts with larger predicted error before
selection.

%% file: chapters/method.tex
\section{Method}\label{sec:method}
We introduce two mechanisms for aligning sparse MoE routing with next-token cross-entropy loss. We first review sparse MoE routing in Section~\ref{subsec:sparse_moe_setting}. Next, Sections~\ref{subsec:token_error_prediction}--\ref{subsec:token_error_supervision} present token-error supervision (TES), and Section~\ref{subsec:affinity_concentration_supervision} introduces affinity-concentration supervision (ACS). Finally, Section~\ref{subsec:training_objective} details the overall training objective and its computational cost.
\subsection{Sparse MoE Routing}\label{subsec:sparse_moe_setting}

Sparse mixture-of-experts models increase parameter capacity while limiting
per-token computation by evaluating only $K$ of $N$ routed experts in each MoE
layer
\citep{shazeer2017outrageously,lepikhin2021gshard,fedus2022switch}.
A learned router assigns each token to its active experts and determines their
contributions to the layer output. Consider a token representation
$\mathbf{h}_t\in\mathbb{R}^{d}$. Omitting the layer index for clarity, the
affinity router computes
\begin{equation}\label{eq:native_affinity}
    \mathbf{a}_t
    =
    \mathbf{W}_{\mathrm{aff}}\mathbf{h}_t
    +\mathbf{b}_{\mathrm{aff}},
    \qquad
    \mathbf{p}_t
    =
    \operatorname{softmax}\!\left(\mathbf{a}_t\right),
    \qquad
    \mathcal{S}_t
    =
    \operatorname{TopK}\!\left(\mathbf{p}_t,K\right).
\end{equation}
Here $\mathbf{W}_{\mathrm{aff}}\in\mathbb{R}^{N\times d}$ and
$\mathbf{b}_{\mathrm{aff}}\in\mathbb{R}^{N}$ are the affinity-router
parameters, while $\mathbf{a}_t,\mathbf{p}_t\in\mathbb{R}^{N}$ are the affinity
logits and probabilities, respectively.
The routed expert branch evaluates the experts in $\mathcal{S}_t$ and combines
their outputs as
\begin{equation}\label{eq:native_moe_output}
    \mathbf{y}_t
    =
    \sum_{i\in\mathcal{S}_t}
    g_{t,i}E_i\!\left(\mathbf{h}_t\right),
\end{equation}
where $g_{t,i}$ denotes the weight derived from
$\mathbf{p}_t$. Depending on the model, these weights may be renormalized over
$\mathcal{S}_t$ or may preserve the selected affinity mass.

The affinity router is learned through the language-model objective and
auxiliary routing objectives, such as load balancing
\citep{shazeer2017outrageously,fedus2022switch}. These objectives make affinity
effective for selecting and combining experts, but do not explicitly supervise
it against the model's token-level error. We therefore retain affinity
as the expert-preference signal and complement it with a separately supervised
token-error signal, introduced in the following subsection.

\subsection{Token-Error Head and Affinity Attenuation}
\label{subsec:token_error_prediction}
\label{subsec:error_aware_attenuation}

TES augments the native router with a lightweight token-error head. From the
shared pre-expert token representation $\mathbf{h}_t$, the head
predicts one positive error score for each expert before expert selection:
\begin{equation}\label{eq:positive_error_head}
    \widehat{\mathbf{e}}_t
    =
    \operatorname{softplus}\!\left(
      \mathbf{W}_{e}\mathbf{h}_t+\mathbf{b}_{e}
    \right),
    \qquad
    \widehat{\mathbf e}_t\in\mathbb R_{>0}^{N}.
\end{equation}
Here $\mathbf{W}_{e}\in\mathbb{R}^{N\times d}$ and
$\mathbf{b}_{e}\in\mathbb{R}^{N}$ are the error-head parameters. The component
$\widehat e_{t,i}$ is the predicted error score for expert $i$ at token $t$;
positivity ensures that the logarithms used below are well defined.

Let $\mathcal{L}_{\mathrm{CE},t}$ denote the realized next-token cross-entropy,
and let $\mathcal S_t^{\mathrm{act}}$ denote the experts executed in the same
forward pass. Thus, $\mathcal S_t^{\mathrm{act}}=\mathcal S_t$ under native
routing and $\mathcal S_t^{\mathrm{act}}=\widetilde{\mathcal S}_t$ under the
error-aware attenuation defined below.
Because this loss evaluates the complete routed prediction rather than an
individual expert, we form a single affinity-weighted prediction over the
active experts:
\begin{equation}\label{eq:aggregate_predicted_error}
    \bar p_{t,i}
    =
    \frac{p_{t,i}}
    {\sum_{j\in\mathcal{S}_t^{\mathrm{act}}}p_{t,j}},
    \quad i\in\mathcal S_t^{\mathrm{act}},
    \qquad
    \widehat e_t
    =
    \sum_{i\in\mathcal{S}_t^{\mathrm{act}}}
    \bar p_{t,i}\widehat e_{t,i}.
\end{equation}

Equation~\ref{eq:aggregate_predicted_error} forms the token-error prediction
supervised in Section~\ref{subsec:token_error_supervision}; active experts with
larger native affinity contribute more to the predicted error.

We use the expert-indexed error predictions to adjust the native affinity
logits $a_{t,i}$, defined in Equation~\ref{eq:native_affinity}, before expert selection:
\begin{equation}\label{eq:attenuated_routing_score}
    \widetilde{a}_{t,i}
    =
    a_{t,i}
    -\gamma\phi_{\tau}(\widehat e_{t,i}),
    \qquad
    \widetilde{\mathbf{p}}_t
    =
    \operatorname{softmax}\!\left(\widetilde{\mathbf{a}}_t\right),
    \qquad
    \widetilde{\mathcal{S}}_t
    =
    \operatorname{TopK}\!\left(\widetilde{\mathbf{p}}_t,K\right).
\end{equation}
Here,
\begin{equation}\label{eq:started_log_transform}
    \phi_{\tau}(\widehat e_{t,i})
    =
    \log\!\left(1+\frac{\widehat e_{t,i}}{\tau}\right),
    \qquad \tau>0,
\end{equation}
is the started logarithm of the predicted error
\citep{rocke2003approximate}. The reference scale $\tau$ controls the
transition between small and large predicted errors, while $\gamma\geq0$
controls the strength of error attenuation relative to native affinity. We set
$\gamma=1$ and $\tau=1$ by default. FFN aggregation uses
$\widetilde{\mathbf p}_t$ under the architecture's native selected-weight
normalization policy; the auxiliary error readout in
Equation~\ref{eq:aggregate_predicted_error} instead uses native affinities.

Exponentiating the adjusted logits factorizes
each softmax term into its native affinity and an inverse-error factor, giving
\begin{equation}\label{eq:attenuated_probability}
    \widetilde p_{t,i}
    =
    \frac{
      p_{t,i}
      \left(1+\widehat e_{t,i}/\tau\right)^{-\gamma}
    }{
      \sum_{j=1}^{N}
      p_{t,j}
      \left(1+\widehat e_{t,j}/\tau\right)^{-\gamma}
    }.
\end{equation}
Because $\phi_{\tau}$ is increasing, experts with larger predicted error
receive lower adjusted logits.
Appendix~\ref{app:shifted_log_geometry} provides the full derivation of
Equation~\ref{eq:attenuated_probability}. Appendix~\ref{app:error_routing_transform_analysis} compares the started-log
transformation with the scale-invariant pure-log alternative.

\subsection{Token-Error Supervision}
\label{subsec:token_error_supervision}

We align $\widehat e_t$ against the realized token loss using the
Itakura--Saito (IS) divergence, a scale-invariant measure of relative
disagreement between positive quantities
\citep{itakura1968analysis,fevotte2009nonnegative}. Because IS requires a
positive observation, we apply a small numerical floor:
\begin{equation}\label{eq:token_is_loss}
    \mathcal{L}^{+}_{\mathrm{CE},t}
    =
    \max\!\left(\mathcal{L}_{\mathrm{CE},t},\varepsilon_{\mathrm{IS}}\right),
    \qquad
    \mathcal{L}_{\mathrm{IS},t}
    =
    \frac{\mathcal{L}^{+}_{\mathrm{CE},t}}
         {\widehat e_t}
    -\log\!\left(
       \frac{\mathcal{L}^{+}_{\mathrm{CE},t}}
            {\widehat e_t}
     \right)-1.
\end{equation}
Exponential negative log-likelihood (ENLL) provides an alternative supervision
objective with the same prediction target but different language-model
gradients. It treats the nonnegative token loss as an observation from an
exponential distribution with conditional mean $\widehat e_t$:
\begin{equation}\label{eq:token_enll_loss}
    \mathcal{L}_{\mathrm{ENLL},t}
    =
    \frac{\mathcal{L}_{\mathrm{CE},t}}
         {\widehat e_t}
         +
         \log \widehat e_t.
\end{equation}

Through the aggregate error prediction in
Equation~\ref{eq:aggregate_predicted_error}, the individual expert scores
receive direct gradients in proportion to their normalized affinities, but are
not identified as counterfactual expert losses. We therefore interpret them as
learned error-aware routing signals. In both TES and ACS, the observed token
CE remains attached to the computation graph, allowing gradients through
both the prediction or routing statistic and the observed loss. Appendices~
\ref{app:aggregate_supervision_expert_scores} and~
\ref{app:attached_token_loss_gradients} analyze their gradient allocation and
the resulting language-model updates.

Both objectives align the aggregate prediction with its supervised token loss.
Across the contextual occurrences $c$ of a fixed target token identity $t'$, the
supervision target is the expected realized token loss. The corresponding
context-averaged token-identity optima are
\begin{equation}\label{eq:token_error_population_optimum}
    \widehat e_{\mathrm{IS},t'}^{\star}=\mathbb E_{c\mid t=t'}[\mathcal L^{+}_{\mathrm{CE},t,c}],\qquad \widehat e_{\mathrm{ENLL},t'}^{\star}=\mathbb E_{c\mid t=t'}[\mathcal L_{\mathrm{CE},t,c}].
\end{equation}
Here, $\widehat e_{\mathrm{IS},t'}^\star$ and
$\widehat e_{\mathrm{ENLL},t'}^\star$ are scalar reference optima obtained by
minimizing the context-averaged supervision loss with respect to a single
prediction for target token identity $t'$. The implemented head remains
context dependent; these reference values do not require identical
predictions across occurrences.
Appendix~\ref{app:token_error_optimality} gives the complete derivation of
Equation~\ref{eq:token_error_population_optimum}.

\subsection{Affinity-Concentration Supervision}
\label{subsec:affinity_concentration_supervision}

Affinity-concentration supervision (ACS) provides an alternative to
the learned token-error head.
We directly align the probabilities by replacing the error score $\widehat e_{t,i}$ in Equation~\ref{eq:aggregate_predicted_error},  with the normalized native affinity $\bar p_{t,i}$:

\begin{equation}\label{eq:affinity_concentration}
    C_t
    =
    \sum_{i\in\mathcal S_t}
    \bar p_{t,i}\bar p_{t,i}
    =
    \sum_{i\in\mathcal S_t}
    \bar p_{t,i}^{2}
    \qquad
    \frac{1}{K}\leq C_t\leq1.
\end{equation}


where $C_t$ denotes the affinity concentration. Its inverse, $1/C_t$, is the Hill effective number of selected
experts, while $-\log C_t$ is the corresponding order-two R\'enyi entropy
\citep{renyi1961measures,hill1973diversity}. This gives $C_t$ a direct routing
interpretation: larger $C_t$ corresponds to a smaller effective number of
experts carrying the selected affinity mass, whereas smaller $C_t$ corresponds
to a more evenly distributed route. In particular, $C_t=1/K$ under uniform
selected affinity and approaches one when affinity concentrates on a single
expert.

ACS applies the token-error supervision objectives introduced in
Section~\ref{subsec:token_error_supervision}, replacing the aggregate prediction
$\widehat e_t$ in
Equations~\ref{eq:token_is_loss} and~\ref{eq:token_enll_loss} with $C_t$.
Here, $C_t$ is a bounded routing statistic, not an unrestricted token-loss
prediction.
Their direct gradients with respect to concentration are
\begin{align}
    \frac{\partial\mathcal L_{\mathrm{ACS\text{-}IS},t}}
         {\partial C_t}
    &=
    \frac{C_t-\mathcal L^{+}_{\mathrm{CE},t}}{C_t^2},
    \label{eq:affinity_concentration_is_gradient}\\
    \frac{\partial\mathcal L_{\mathrm{ACS\text{-}ENLL},t}}
         {\partial C_t}
    &=
    \frac{C_t-\mathcal L_{\mathrm{CE},t}}{C_t^2}.
    \label{eq:affinity_concentration_enll_gradient}
\end{align}
When the IS floor is inactive, the two objectives produce the same direct
concentration gradient. This gradient favors sharper selected affinity when
token loss exceeds concentration and flatter affinity otherwise; the complete
parameter update also depends on other gradient paths.
ACS therefore uses token error to supervise routing affinities, without adding an error head or applying an
error-aware routing transformation at inference.
Appendix~\ref{app:affinity_softmax_error_supervision} gives its bounded optimum,
router-logit gradients, and additional analysis.

\subsection{Overall Training Objective and Computational Scope}
\label{subsec:training_objective}

Restoring the layer indices omitted above, let $\mathcal{B}_{\mathrm{tok}}$
denote the supervised token positions in a minibatch, excluding padding and
ignored targets, and let $\mathcal{M}$ denote the MoE layers receiving
supervision. Let
$m\in\{\mathrm{TES\text{-}IS},\mathrm{TES\text{-}ENLL},
\mathrm{ACS\text{-}IS},\mathrm{ACS\text{-}ENLL}\}$ denote the active
supervision objective, and let $\mathcal L_{\mathrm{CE}}$ denote the
cross-entropy averaged over $\mathcal B_{\mathrm{tok}}$. We average the active
objective over supervised tokens and layers as
\begin{equation}\label{eq:averaged_training_losses}
    \mathcal{L}_m
    =
    \frac{1}
         {|\mathcal{M}|
          |\mathcal{B}_{\mathrm{tok}}|}
    \sum_{l\in\mathcal{M}}
    \sum_{t\in\mathcal{B}_{\mathrm{tok}}}
    \mathcal{L}_{m,t}^{(l)}.
\end{equation}
Here $\mathcal{L}_{m,t}^{(l)}$ is evaluated using the learned aggregate
prediction $\widehat e_t^{(l)}$ when $m$ is a TES objective, or the affinity
concentration $C_t^{(l)}$ when $m$ is an ACS objective. The complete training
objective is
\begin{equation}\label{eq:total_loss}
    \mathcal{L}_{\mathrm{total}}
    =
    \mathcal{L}_{\mathrm{CE}}
    +
    \lambda_m\mathcal{L}_m
    +
    \lambda_{\mathrm{aux}}\mathcal{L}_{\mathrm{aux}}.
\end{equation}
Here, $\mathcal{L}_{\mathrm{aux}}$ is the architecture-native router regularizer,
typically encouraging balanced expert usage
\citep{shazeer2017outrageously,fedus2022switch}; $\lambda_m,\lambda_{\mathrm{aux}}
\geq0$ are regularization weights of $\mathcal{L}_m$ and $\mathcal{L}_{\mathrm{aux}}$, respectively.
Each MoE layer equipped with TES includes an additional linear projection with
$N(d+1)$ parameters and $\mathcal{O}(Nd)$ operations per token, while preserving
the native top-$K$ execution budget. ACS adds no parameters and retains native
routing at inference.
For MCQA adaptation, the primary task term is answer-choice CE, while the
auxiliary target remains next-token CE (Appendix~\ref{app:experimental_configuration}).

%% file: chapters/results.tex
\section{Experiments and Results}\label{sec:results}
We evaluate whether aligning MoE routing with token loss improves downstream accuracy
relative to parameter-matched controls under a common supervision setting
(Section~\ref{subsec:accuracy_checkpoint_selection}). We then analyze whether predicted
error tracks observed cross-entropy loss (Section~\ref{subsec:error_prediction_analysis}).
Finally, we ablate the number of supervised MoE layers
(Section~\ref{subsec:supervision_depth}).
\subsection{Experimental Setting}\label{subsec:experimental_setting}
We evaluate Granite 3.1 3B-A800M and OLMoE-1B-7B-SFT
\citep{ibm2024granite31,muennighoff2024olmoe} on ARC-Challenge, OpenBookQA,
SciQ, and MedMCQA \citep{clark2018arc,mihaylov2018openbookqa,welbl2017sciq,pal2022medmcqa}.
Within each backbone and dataset, all
configurations use the same registered train, validation, and test splits and
the same answer-text multiple-choice protocol without in-context examples.
Candidate answers are
ranked by their mean conditional token log-probability. We report accuracy on a
fixed 500-example test set; split construction and prompting are documented in
Appendix~\ref{app:experimental_configuration}.

Where applicable, we follow VMoER's Stage-1 MAP adaptation setting
\citep{li2026vmoer}: the same dataset suite and target split sizes,
three-epoch LoRA adaptation of attention Q/K/V and routed experts, AdamW, and
the Granite learning rate, schedule, warmup, and effective batch size. Because
VMoER does not release split identities, we construct deterministic partitions
targeting the reported sizes and filter invalid examples. Our answer-text candidate scoring differs from
VMoER's generated-letter protocol, and the OLMoE recipe is adapted to that
backbone.

Both models are adapted for three epochs using rank-8 LoRA with $\alpha=8$ and
dropout $0.05$. Granite uses a learning rate of $10^{-4}$ and effective batch
size 16, whereas OLMoE uses $2\times10^{-5}$ and batch size 8. We evaluate
training and initialization seeds $42$, $43$, and $45$ for the primary
accuracy comparison. Method-specific
supervision is applied at the final sparse MoE layer in the primary comparison.
Native affinity parameters remain frozen except in rows explicitly marked
``+ affinity tuning,'' and every configuration preserves the native
expert-execution budget.

For a fair evaluation, we compare ACS and TES with parameter-matched baselines
under the IS and ENLL alignment schemes. ACS is compared with the CE fine-tuning
baseline, while TES is compared with Dual Affinity, which adds an affinity head
with the same parameter count and averages the two router outputs. To match the
TES configuration, the original pretrained router is kept frozen and only the
second affinity head is updated during fine-tuning. For each configuration, we
report the mean and standard deviation across three seeds. Unless stated
otherwise, both methods use a supervision coefficient of $\lambda_m=10^{-3}$.
Validation-NLL-selected
checkpoints and complete training configurations are described in
Appendix~\ref{app:experimental_configuration}. TES sensitivity to the
error-head learning rate and attenuation scale is reported in
Appendix~\ref{app:tes_hyperparameter_sensitivity}.

\subsection{Downstream Accuracy}
\label{subsec:accuracy_checkpoint_selection}

\begin{table}[t]
        \centering
    \scriptsize
    \caption{Test answer-choice accuracy (\%) for Granite 3.1 3B-A800M and
    OLMoE-1B-7B-SFT across four MCQA benchmarks. Entries are mean $\pm$ sample
    standard deviation over seeds 42, 43, and 45. Within each backbone, rows
    form three parameter-matched comparison groups: CE versus ACS with frozen
    affinity, CE versus ACS with affinity tuning, and Dual Affinity versus TES.
    All entries use epoch-3 checkpoints except the OLMoE affinity-tuned group,
    which uses validation-NLL-selected checkpoints. Bold marks the highest mean
    within each backbone, benchmark, and comparison group.
    $^{\dagger}$ SciQ is near saturation.}
    \label{tab:downstream_accuracy}

    \begin{tabular}{@{}llcccc@{}}
        \toprule
        Model & Method & ARC-Challenge & OpenBookQA & SciQ$^{\dagger}$ & MedMCQA \\
        \midrule
        Granite 3.1 & CE
            & $65.33\pm2.01$ & $71.67\pm1.01$ & $88.40\pm0.53$
            & $39.60\pm9.57$ \\
        & ACS-ENLL (Ours)
            & $67.40\pm0.00$ & $72.60\pm0.53$ & $89.27\pm0.81$
            & $42.80\pm0.40$ \\
        & ACS-IS (Ours)
            & $\mathbf{67.87\pm0.76}$ & $\mathbf{72.73\pm0.95}$
            & $\mathbf{89.87\pm0.70}$ & $\mathbf{42.87\pm0.76}$ \\
        \cmidrule(lr){2-6}
        & CE + affinity tuning
            & $67.53\pm1.14$ & $\mathbf{74.00\pm0.72}$ & $88.60\pm0.40$
            & $41.73\pm1.22$ \\
        & ACS-ENLL + affinity tuning (Ours)
            & $69.87\pm1.33$ & $73.53\pm0.90$
            & $89.20\pm0.92$ & $\mathbf{42.13\pm1.27}$ \\
        & ACS-IS + affinity tuning (Ours)
            & $\mathbf{70.60\pm1.80}$ & $73.93\pm0.31$
            & $\mathbf{89.53\pm0.42}$ & $40.67\pm1.45$ \\
        \cmidrule(lr){2-6}
        & Dual Affinity
            & $65.33\pm2.00$ & $71.27\pm1.86$ & $88.27\pm0.31$
            & $37.40\pm7.79$ \\
        & TES-ENLL (Ours)
            & $65.53\pm0.42$ & $72.27\pm0.31$ & $\mathbf{88.53\pm1.67}$
            & $\mathbf{43.67\pm1.81}$ \\
        & TES-IS (Ours)
            & $\mathbf{66.40\pm0.35}$ & $\mathbf{73.00\pm0.53}$ & $88.33\pm0.64$
            & $\mathbf{43.67\pm1.10}$ \\
        \midrule
        OLMoE & CE
            & $\mathbf{63.27\pm0.31}$ & $\mathbf{70.40\pm0.40}$ & $88.80\pm0.40$
            & $\mathbf{44.93\pm0.70}$ \\
        & ACS-ENLL (Ours)
            & $62.93\pm0.64$ & $70.20\pm0.35$ & $88.73\pm0.23$
            & $44.53\pm0.76$ \\
        & ACS-IS (Ours)
            & $62.73\pm0.31$ & $69.93\pm0.46$ & $\mathbf{88.87\pm0.12}$
            & $44.20\pm0.53$ \\
        \cmidrule(lr){2-6}
        & CE + affinity tuning
            & $63.53\pm0.58$ & $\mathbf{71.87\pm0.42}$
            & $88.80\pm0.20$ & $44.47\pm0.46$ \\
        & ACS-ENLL + affinity tuning (Ours)
            & $\mathbf{63.80\pm0.53}$ & $71.00\pm0.60$
            & $88.87\pm0.42$ & $44.27\pm0.76$ \\
        & ACS-IS + affinity tuning (Ours)
            & $63.60\pm0.35$ & $69.33\pm2.42$
            & $\mathbf{89.07\pm0.31}$ & $\mathbf{44.53\pm1.17}$ \\
        \cmidrule(lr){2-6}
        & Dual Affinity
            & $61.27\pm1.40$ & $70.80\pm1.06$ & $\mathbf{89.80\pm0.69}$
            & $43.93\pm0.76$ \\
        & TES-ENLL (Ours)
            & $62.93\pm0.12$ & $69.87\pm0.61$ & $88.80\pm0.35$
            & $43.87\pm0.81$ \\
        & TES-IS (Ours)
            & $\mathbf{63.53\pm0.46}$ & $\mathbf{70.93\pm0.23}$
            & $88.80\pm0.40$ & $\mathbf{44.53\pm0.12}$ \\
        \bottomrule
    \end{tabular}
\end{table}

Table~\ref{tab:downstream_accuracy} compares TES and ACS across four datasets and two models.
TES-IS improves ARC-Challenge accuracy over Dual Affinity by 1.07 percentage
points on Granite and 2.26 points on OLMoE. It also improves OpenBookQA by
1.73 and 0.13 points, respectively, and OLMoE MedMCQA by 0.60 points.
TES-ENLL improves ARC-Challenge on both backbones, although by smaller
amounts. The objective ordering varies across tasks:
on Granite SciQ, ENLL gives the larger TES gain of 0.26 points, whereas
both TES objectives fall 1.00 point below Dual Affinity on OLMoE SciQ.

ACS produces its clearest gains on Granite. With native affinity frozen,
ACS-IS improves over CE by 2.54 points on ARC-Challenge, 1.06 on OpenBookQA,
and 1.47 on SciQ. Allowing affinity tuning changes the reference comparison:
against affinity-tuned CE, ACS-IS improves ARC-Challenge by 3.07 points and
SciQ by 0.93 points, while ACS-ENLL improves MedMCQA by 0.40 points.
On OLMoE, ACS gains are smaller and task-dependent, and neither objective
improves OpenBookQA in either affinity setting. These results distinguish
the benefit of concentration supervision from that of simply unfreezing
the router.

\begin{table}[t]
    \scriptsize
    \caption{Effect of increasing the method-supervision coefficient to
    $\lambda_m=10^{-2}$. Each cell reports epoch-3 test accuracy (\%) followed
    in brackets by its change from the corresponding $\lambda_m=10^{-3}$
    method result, in percentage points. Bold indicates higher mean accuracy
    than the corresponding $\lambda_m=10^{-3}$ configuration.}
    \label{tab:method_coefficient_impact}
    \centering
    \begin{tabular}{@{}lllcccc@{}}
        \toprule
        Model & Mechanism & Objective
            & ARC-C & OpenBookQA & SciQ & MedMCQA \\
        \midrule
        Granite & TES, frozen & IS
            & $\mathbf{68.27}\,[+1.87]$
            & $72.47\,[-0.53]$
            & $\mathbf{89.20}\,[+0.87]$
            & $43.67\,[+0.00]$ \\
        & & ENLL
            & $\mathbf{67.73}\,[+2.20]$
            & $\mathbf{72.73}\,[+0.47]$
            & $\mathbf{90.13}\,[+1.60]$
            & $43.40\,[-0.27]$ \\
        \midrule
        Granite & ACS, trainable & IS
            & $\mathbf{71.20}\,[+0.60]$
            & $\mathbf{74.53}\,[+0.60]$
            & $88.80\,[-0.73]$
            & $\mathbf{41.07}\,[+0.40]$ \\
        & & ENLL
            & $\mathbf{70.47}\,[+0.60]$
            & $\mathbf{75.20}\,[+1.67]$
            & $89.00\,[-0.20]$
            & $41.47\,[-0.66]$ \\
        \midrule
        OLMoE & ACS, frozen & IS
            & $\mathbf{63.40}\,[+0.67]$
            & $69.80\,[-0.13]$
            & $\mathbf{89.67}\,[+0.80]$
            & $44.07\,[-0.13]$ \\
        & & ENLL
            & $\mathbf{63.93}\,[+1.00]$
            & $69.20\,[-1.00]$
            & $\mathbf{90.07}\,[+1.34]$
            & $44.13\,[-0.40]$ \\
        \bottomrule
    \end{tabular}%
\end{table}

Table~\ref{tab:method_coefficient_impact} evaluates stronger supervision by
increasing $\lambda_m$ to $10^{-2}$, reporting both accuracy and changes from
the corresponding result in Table~\ref{tab:downstream_accuracy}.
Stronger supervision
improves accuracy on ARC-Challenge. Granite TES-IS reaches 68.27\%, an increase of
1.87 points over the corresponding $10^{-3}$ result and 2.94 points over Dual Affinity.
Granite TES-ENLL reaches 90.13\% on SciQ, improving by 1.60 points over its
smaller-coefficient result and by 1.86 points over its control. Stronger
supervision also benefits ACS. Granite affinity-tuned ACS-ENLL improves
OpenBookQA by 1.67 points relative to its $10^{-3}$ setting.

These additional gains do not imply that the larger coefficient is preferable
for every task. For example, Granite TES-IS loses 0.53 points on OpenBookQA,
and both Granite affinity-tuned ACS objectives lose accuracy on SciQ.
We therefore retain the common coefficient in
Table~\ref{tab:downstream_accuracy} and report the complete paired changes
in Table~\ref{tab:method_coefficient_impact}, rather than selecting the better
coefficient separately for each test benchmark. Together, the comparisons
show accuracy improvements under a shared setting and further gains from
stronger supervision on several tasks.

\subsection{Error-Prediction Analysis}
\label{subsec:error_prediction_analysis}

\begin{figure}[t]
    \centering
    \begin{minipage}[t]{0.495\linewidth}
        \centering
        \includegraphics[width=\linewidth]{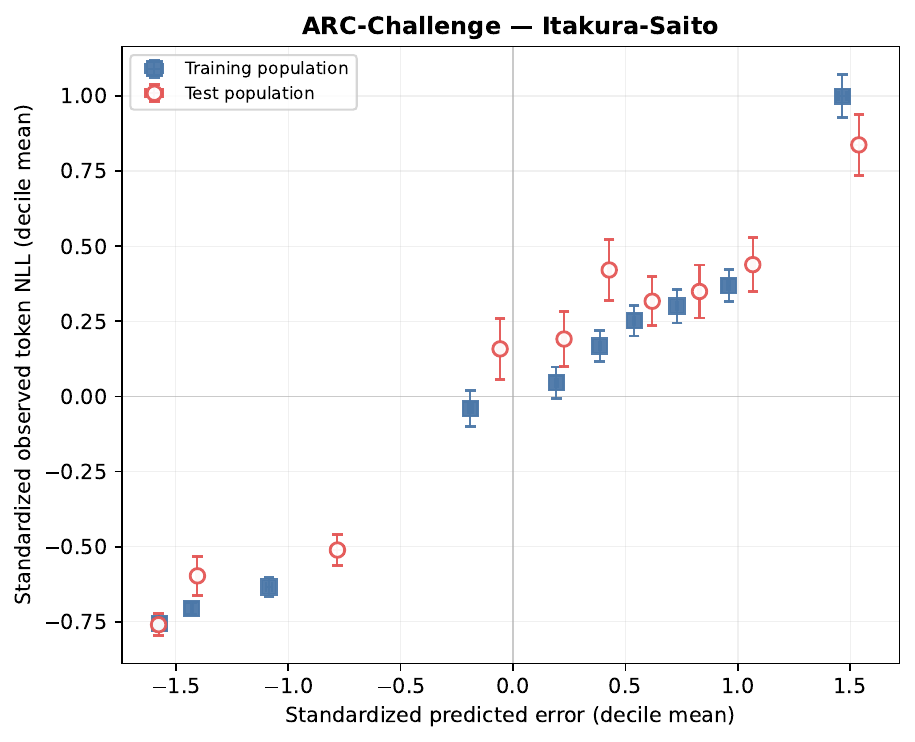}
        \vspace{-0.8em}
        \textbf{(a) Itakura--Saito}
    \end{minipage}\hfill
    \begin{minipage}[t]{0.495\linewidth}
        \centering
        \includegraphics[width=\linewidth]{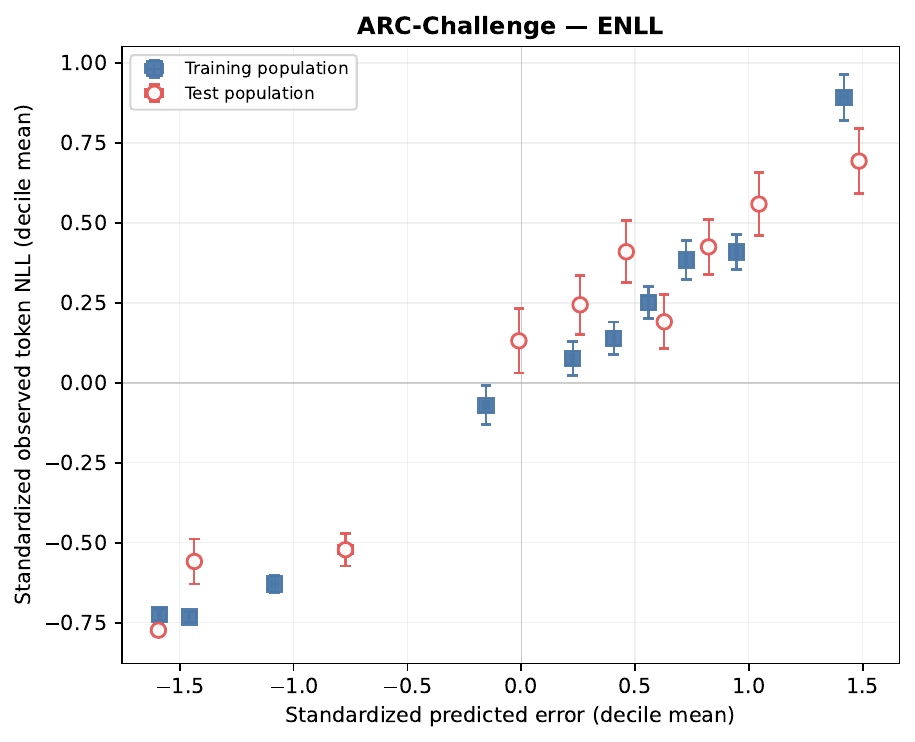}
        \vspace{-0.8em}
        \textbf{(b) ENLL}
    \end{minipage}
    \caption{Aggregate token-error prediction on ARC-Challenge for Granite TES
    with started-log attenuation. The horizontal and vertical axes show
    standardized predicted error and observed token NLL, respectively; both use
    training-set moments. Points are equal-count decile means over
    layer--token-identity groups after averaging contextual occurrences within each
    group and then averaging group means over seeds 42--44. Whiskers show
    $\pm1$ standard error across groups. Blue squares and red circles denote the
    training and test populations, respectively.}
    \label{fig:arc_error_prediction}
\end{figure}

Figure~\ref{fig:arc_error_prediction} examines whether the aggregate prediction
$\widehat e_t$ correlates with observed next-token NLL across
layer--token-identity groups, after averaging contextual occurrences within
each group. Each group is defined by a routed layer $\ell$ and target token
identity $t'$. Within each seed, we average predicted and observed errors over
the contextual sequences whose next-token target is $t'$. We then average
the group means across seeds and
form ten equal-count bins by predicted error. This is the layer-specific
empirical counterpart of the context averaging in Equation~\ref{eq:token_error_population_optimum}.
In the figure, we standardize both quantities. 

Both objectives capture token difficulty. Across test-set layer--token-identity
groups, Pearson/Spearman
correlations are $0.452/0.564$ for IS and $0.449/0.562$ for ENLL.
Observed NLL increases across eight of the nine adjacent-decile transitions
for both objectives. These standardized plots assess association and
out-of-sample shift, rather than absolute NLL calibration.
Appendix~\ref{app:cross_dataset_error_prediction} presents the corresponding
OpenBookQA, SciQ, and MedMCQA analyses.

\subsection{Ablation Studies}
\label{subsec:supervision_depth}
Table~\ref{tab:supervision_depth} compares the primary final-layer setting with
all-layer ACS on OLMoE while keeping native affinity frozen and the supervision
coefficient fixed. Extending ACS to all 16 MoE layers changes accuracy by at
most $0.46$ percentage points across objectives and datasets, with neither
scope consistently performing better; final-layer supervision therefore
remains the simpler competitive default. Additional ablations of the
error-to-routing transformation, TES hyperparameters, supervision strength,
affinity-router fine-tuning across layer scopes, and depth-dependent
coefficient scaling are provided in
Appendices~\ref{app:error_routing_transform_analysis},
\ref{app:tes_hyperparameter_sensitivity},
\ref{app:acs_coefficient_sensitivity},
\ref{app:affinity_router_layer_ablation}, and
\ref{app:acs_depth_coefficient_scaling}.

\begin{table}[t]
    \caption{OLMoE epoch-3 test accuracy (\%; mean $\pm$ sample standard
    deviation over seeds 42, 43, and 45) for final-layer and all-layer ACS with
    frozen native affinity and $\lambda_m=10^{-3}$. Bold marks the higher mean
    within each objective and benchmark.}
    \label{tab:supervision_depth}
    \centering
    \scriptsize
    \begin{tabular}{@{}llcccc@{}}
        \toprule
        Objective & Supervised MoE layers & ARC-Challenge & OpenBookQA & SciQ & MedMCQA \\
        \midrule
        ACS-ENLL & Final layer
            & $62.93\pm0.64$ & $\mathbf{70.20\pm0.35}$ & $88.73\pm0.23$
            & $\mathbf{44.53\pm0.76}$ \\
        ACS-ENLL & All 16 layers
            & $\mathbf{63.13\pm0.42}$ & $69.80\pm0.40$
            & $\mathbf{88.80\pm0.40}$ & $44.07\pm0.81$ \\
        \cmidrule(lr){1-6}
        ACS-IS & Final layer
            & $62.73\pm0.31$ & $69.93\pm0.46$ & $88.87\pm0.12$
            & $\mathbf{44.20\pm0.53}$ \\
        ACS-IS & All 16 layers
            & $\mathbf{63.07\pm0.50}$ & $\mathbf{70.13\pm0.12}$
            & $\mathbf{89.07\pm0.31}$ & $44.00\pm0.35$ \\
        \bottomrule
    \end{tabular}
\end{table}

%% file: chapters/conclusion.tex
\section{Conclusion}\label{sec:conclusion}
\textbf{Summary.} We presented two complementary mechanisms for aligning sparse MoE routing
with token-level loss. TES predicts expert-error scores that guide
affinity attenuation, while ACS directly aligns native affinity
concentration without an additional head. Across two MoE backbones and four MCQA benchmarks, both
mechanisms yield accuracy gains over their corresponding controls while
preserving the native sparse execution budget and aggregation policy.
Stronger supervision provides additional gains, with TES-IS reaching a
2.94-percentage-point improvement over Dual Affinity on Granite
ARC-Challenge. These findings establish direct token-loss supervision as a
useful complement to affinity-based routing and highlight supervision
strength and depth as important design choices.

\textbf{Limitations and Future Work.} Our evaluation covers two sparse MoE backbones, four multiple-choice
question-answering benchmarks, fixed data splits, and three training seeds.
Larger-scale models and pretraining-scale optimization are outside the scope
of this evaluation. Finally, although both
methods preserve the native top-$K$ expert budget, we do not measure end-to-end
latency or memory overhead from the additional TES projection. Future work will extend our approach to pretraining paradigms and to additional backbones and datasets.

%% file: chapters/appendix.tex
\section{Appendix}\label{sec:appendix}

\subsection{Scalar Loss Geometry and Token-Identity Prediction Optima}
\label{app:token_error_optimality}

For this analysis, let $t$ denote the target token identity of an occurrence,
let $t'$ denote a fixed target token identity, and let $c$ index a supervised
prediction context whose next-token target is $t$. The error head
receives the corresponding hidden representation $\mathbf h_{t,c}$ at the
prediction position. Together with the selected native-affinity weights,
its expert scores form the positive aggregate $\widehat e_{t,c}$ in
Equation~\ref{eq:aggregate_predicted_error}. Let
$\mathcal L_{\mathrm{CE},t,c}$ denote the realized next-token loss for that
occurrence. We omit the layer index and initially treat each prediction as an
independently adjustable scalar while holding the observed losses fixed.
The pair $(t,c)$ identifies one occurrence; elsewhere, the main text uses
$t$ alone as a token-position index. Expectations below are empirical averages
or population expectations with finite mean token loss.

\paragraph{Scalar loss geometry.}
For one contextual occurrence, differentiation with respect to the aggregate
prediction gives
\begin{align}
    \frac{\partial\mathcal L_{\mathrm{IS},t,c}}
         {\partial\widehat e_{t,c}}
    &=
    \frac{\widehat e_{t,c}-\mathcal L^{+}_{\mathrm{CE},t,c}}
         {\widehat e_{t,c}^{2}},
    \label{eq:appendix_is_direct_gradient}\\
    \frac{\partial\mathcal L_{\mathrm{ENLL},t,c}}
         {\partial\widehat e_{t,c}}
    &=
    \frac{\widehat e_{t,c}-\mathcal L_{\mathrm{CE},t,c}}
         {\widehat e_{t,c}^{2}}.
    \label{eq:appendix_enll_direct_gradient}
\end{align}
For a positive target, the derivative is negative below the target and positive
above it. Thus, an independently adjustable occurrence-level prediction is
minimized at its realized target. If an ENLL target is zero, its objective is
$\log\widehat e_{t,c}$ and tends to $-\infty$ as
$\widehat e_{t,c}\rightarrow0$; the ENLL statements below assume positive
targets for the occurrence-level optimum. A positive mean target suffices
for the context-averaged ENLL optimum below.

\paragraph{Token-identity optimum.}
For a fixed target token identity $t'$, consider a positive scalar reference
$\widehat e_{t'}$, held constant inside the context average. This reference
summarizes the identity's supervision target; it is not defined as the average
output of the implemented contextual head. Its objectives are
\begin{align}
    \overline{\mathcal L}_{\mathrm{IS},t'}(\widehat e_{t'})
    &:={}
    \mathbb E_{c\mid t=t'}\!\left[
      \frac{\mathcal L^{+}_{\mathrm{CE},t,c}}{\widehat e_{t'}}
      -\log\!\left(
        \frac{\mathcal L^{+}_{\mathrm{CE},t,c}}{\widehat e_{t'}}
      \right)-1
    \right],
    \label{eq:appendix_token_family_is}\\
    \overline{\mathcal L}_{\mathrm{ENLL},t'}(\widehat e_{t'})
    &:={}
    \mathbb E_{c\mid t=t'}\!\left[
      \frac{\mathcal L_{\mathrm{CE},t,c}}{\widehat e_{t'}}
      +\log\widehat e_{t'}
    \right].
    \label{eq:appendix_token_family_enll}
\end{align}
The overline denotes averaging over contexts whose target token identity is $t'$.
In a finite corpus, this expectation is the arithmetic mean
over those occurrences, so frequent contexts contribute according to their
empirical frequency. Differentiation yields
\begin{align}
    \frac{\partial\overline{\mathcal L}_{\mathrm{IS},t'}}
         {\partial\widehat e_{t'}}
    &=
    \frac{
      \widehat e_{t'}-
      \mathbb E_{c\mid t=t'}[\mathcal L^{+}_{\mathrm{CE},t,c}]
    }{\widehat e_{t'}^{2}},
    \label{eq:appendix_token_family_is_gradient}\\
    \frac{\partial\overline{\mathcal L}_{\mathrm{ENLL},t'}}
         {\partial\widehat e_{t'}}
    &=
    \frac{
      \widehat e_{t'}-
      \mathbb E_{c\mid t=t'}[\mathcal L_{\mathrm{CE},t,c}]
    }{\widehat e_{t'}^{2}}.
    \label{eq:appendix_token_family_enll_gradient}
\end{align}
Therefore, the unique positive token-identity optima are
\begin{equation}\label{eq:appendix_token_family_optima}
    \widehat e_{\mathrm{IS},t'}^{\star}
    =\mathbb E_{c\mid t=t'}[\mathcal L^{+}_{\mathrm{CE},t,c}],
    \qquad
    \widehat e_{\mathrm{ENLL},t'}^{\star}
    =\mathbb E_{c\mid t=t'}[\mathcal L_{\mathrm{CE},t,c}].
\end{equation}
The derivatives change sign from negative to positive at these values, proving
global optimality on the positive scalar domain. The IS floor guarantees a
positive IS optimum; ENLL requires a positive conditional mean.
This establishes Equation~\ref{eq:token_error_population_optimum} as a
token-identity scalar reference. Away from the numerical IS floor, IS and ENLL
have the same reference target.

\paragraph{Shared-head interpretation.}
The implemented prediction $\widehat e_{t,c}$ can vary across contexts through
$\mathbf h_{t,c}$, the affinity weights, and the active expert set.
Its direct parameter gradient averages the scalar derivatives above multiplied
by the corresponding prediction gradients. Parameter sharing therefore does
not guarantee either occurrence-level loss matching or equality between the
average prediction and average loss within each target identity. In particular,
averaging contextual predictions before applying a loss is a different
objective from averaging their individual supervision losses.
For an unrestricted predictor, the corresponding expected-loss optimum
conditions on the information available to that predictor; the next-token
target identity is not itself an input to the error head.
Equation~\ref{eq:token_error_population_optimum} is consequently a reference
for grouped evaluation, not a calibration guarantee for the trained head.
During joint training, routing and model updates also change the observed
losses; Appendices~\ref{app:aggregate_supervision_expert_scores}
and~\ref{app:attached_token_loss_gradients} analyze these gradient paths.

\subsection{Gradient Allocation and Expert-Score Identifiability}
\label{app:aggregate_supervision_expert_scores}

We now return to the main-text convention in which $t$ indexes a supervised
token position.
Equation~\ref{eq:aggregate_predicted_error} supervises one aggregate prediction
even though affinity attenuation uses all $N$ expert-indexed scores and the
aggregate reads only the $K$ executed experts. We first
characterize how the aggregate objective distributes its direct gradient and
then clarify which properties of the individual scores this supervision can
identify.

Holding the active set, normalized affinity weights, and observed loss fixed,
\begin{equation}\label{eq:appendix_aggregate_component_derivative}
    \frac{\partial\widehat e_t}
         {\partial\widehat e_{t,i}}
    =
    \bar p_{t,i},
    \qquad i\in\mathcal S_t^{\mathrm{act}}.
\end{equation}
Applying the chain rule to the two token-error objectives gives
\begin{align}
    \frac{\partial\mathcal{L}_{\mathrm{IS},t}}
         {\partial\widehat e_{t,i}}
    &={}
    \bar p_{t,i}
    \frac{
      \widehat e_t-\mathcal{L}^{+}_{\mathrm{CE},t}
    }{
      \widehat e_t^2
    },
    \label{eq:appendix_is_component_gradient}\\
    \frac{\partial\mathcal{L}_{\mathrm{ENLL},t}}
         {\partial\widehat e_{t,i}}
    &={}
    \bar p_{t,i}
    \frac{
      \widehat e_t-\mathcal{L}_{\mathrm{CE},t}
    }{
      \widehat e_t^2
    }.
    \label{eq:appendix_enll_component_gradient}
\end{align}
Because $\bar p_{t,i}\geq 0$, every active expert score receives a gradient with
the same sign as the aggregate prediction gradient. Its magnitude is scaled by
$\bar p_{t,i}$: among experts receiving the same aggregate error signal,
experts with larger native affinity receive proportionally larger direct
score gradients. Parameter updates also depend on the head's Jacobian and
the optimizer.

This gradient allocation does not make the individual scores independently
identifiable from one fixed aggregate. For perturbations $\Delta_{t,i}$ small
enough to preserve positive scores and the active set,
\begin{equation}\label{eq:appendix_zero_aggregate_perturbation}
    \sum_{i\in\mathcal{S}_t^{\mathrm{act}}}
    \bar p_{t,i}\Delta_{t,i}
    =0
    \quad\Longrightarrow\quad
    \sum_{i\in\mathcal{S}_t^{\mathrm{act}}}
    \bar p_{t,i}(\widehat e_{t,i}+\Delta_{t,i})
    =
    \widehat e_t.
\end{equation}
For $K>1$, this single scalar constraint leaves a local $(K-1)$-dimensional
family of active-score perturbations. This is an occurrence-level property
of the readout, not a proof of parameter non-identifiability across the
dataset. Such perturbations can change attenuated aggregation weights and
the resulting CE even when the expert set stays fixed, so they need not
preserve the complete training loss. The aggregate target alone does not
identify $\widehat e_{t,i}$ as the counterfactual loss obtained by executing
expert $i$ alone.

Differences among expert-indexed scores can nevertheless emerge because the
error head is shared across tokens with different representations, affinities,
and active sets. We therefore interpret their relative values as learned
error-aware routing signals and evaluate them through affinity attenuation,
without claiming independently calibrated expert losses. The displayed
derivatives isolate the direct path through the aggregate readout; full
parameter gradients may additionally pass through $\bar p_{t,i}$,
$\mathbf h_t$, and the realized CE via attenuation. Unselected scores have
zero direct readout gradient but may receive gradients through the routed
model where its normalization policy permits. All local derivatives hold
away from changes in the discrete top-$K$ set.

\subsection{Token-Loss Gradient Analysis}
\label{app:attached_token_loss_gradients}

For clarity, we omit the numerical IS floor and take
$\mathcal{L}_{\mathrm{CE},t}>0$. Although IS and ENLL have the same optimum for
$\widehat e_t$, their derivatives with respect to the observed token loss
differ. For IS,
\begin{align}
    \frac{\partial\mathcal{L}_{\mathrm{IS},t}}
         {\partial\mathcal{L}_{\mathrm{CE},t}}
    &={}
    \frac{\partial}{\partial\mathcal{L}_{\mathrm{CE},t}}
    \left[
      \frac{\mathcal{L}_{\mathrm{CE},t}}{\widehat e_t}
      -\log\!\left(
        \frac{\mathcal{L}_{\mathrm{CE},t}}{\widehat e_t}
      \right)-1
    \right]
    \notag\\
    &={}
    \frac{1}{\widehat e_t}
    -\frac{1}{\mathcal{L}_{\mathrm{CE},t}}.
    \label{eq:appendix_is_token_loss_gradient}
\end{align}
For ENLL,
\begin{align}
    \frac{\partial\mathcal{L}_{\mathrm{ENLL},t}}
         {\partial\mathcal{L}_{\mathrm{CE},t}}
    &={}
    \frac{\partial}{\partial\mathcal{L}_{\mathrm{CE},t}}
    \left[
      \log\widehat e_t
      +\frac{\mathcal{L}_{\mathrm{CE},t}}{\widehat e_t}
    \right]
    \notag\\
    &={}
    \frac{1}{\widehat e_t}.
    \label{eq:appendix_enll_token_loss_gradient}
\end{align}

Let $\theta$ denote any trainable parameters of the routed model, including
the error head when applicable. Away from top-$K$ boundaries, applying the chain
rule to the complete auxiliary objectives gives
\begin{align}
    \nabla_{\theta}\mathcal{L}_{\mathrm{IS},t}
    &={}
    \underbrace{
    \left(
      \frac{1}{\widehat e_t}
      -\frac{1}{\mathcal{L}_{\mathrm{CE},t}}
    \right)
    \nabla_{\theta}\mathcal{L}_{\mathrm{CE},t}
    }_{\text{gradient through the token loss}}
    +
    \underbrace{
    \frac{\widehat e_t-\mathcal{L}_{\mathrm{CE},t}}
         {\widehat e_t^2}
    \nabla_{\theta}\widehat e_t
    }_{\text{gradient through the error prediction}},
    \label{eq:appendix_is_complete_gradient}\\
    \nabla_{\theta}\mathcal{L}_{\mathrm{ENLL},t}
    &={}
    \underbrace{
    \frac{1}{\widehat e_t}
    \nabla_{\theta}\mathcal{L}_{\mathrm{CE},t}
    }_{\text{gradient through the token loss}}
    +
    \underbrace{
    \frac{\widehat e_t-\mathcal{L}_{\mathrm{CE},t}}
         {\widehat e_t^2}
    \nabla_{\theta}\widehat e_t
    }_{\text{gradient through the error prediction}}.
    \label{eq:appendix_enll_complete_gradient}
\end{align}

The second term is identical for IS and ENLL and trains the aggregate
prediction toward the observed token loss. The difference lies in the first
term, which changes how each auxiliary objective contributes to the model's
task-loss gradient. At the matched prediction
$\widehat e_t=\mathcal{L}_{\mathrm{CE},t}$, the auxiliary gradients reduce to
\begin{equation}\label{eq:appendix_matched_token_loss_gradients}
    \left.
    \nabla_{\theta}\mathcal{L}_{\mathrm{IS},t}
    \right|_{\widehat e_t=\mathcal{L}_{\mathrm{CE},t}}
    =0,
    \qquad
    \left.
    \nabla_{\theta}\mathcal{L}_{\mathrm{ENLL},t}
    \right|_{\widehat e_t=\mathcal{L}_{\mathrm{CE},t}}
    =
    \frac{1}{\mathcal{L}_{\mathrm{CE},t}}
    \nabla_{\theta}\mathcal{L}_{\mathrm{CE},t}.
\end{equation}
For a next-token CE primary objective, we isolate the corresponding per-token
terms from Equation~\ref{eq:total_loss}, suppressing token and layer averages
and unrelated auxiliary losses. These combined-gradient identities concern
that objective, not the answer-choice CE task term used in our MCQA
experiments; the auxiliary-gradient identities above still apply.
At the matched prediction,
\begin{equation}\label{eq:appendix_matched_total_model_gradients}
    \begin{aligned}
        \left.
        \nabla_{\theta}
        \left(
          \mathcal{L}_{\mathrm{CE},t}
          +\lambda_m\mathcal{L}_{\mathrm{IS},t}
        \right)
        \right|_{\widehat e_t=\mathcal{L}_{\mathrm{CE},t}}
        &={}
        \nabla_{\theta}\mathcal{L}_{\mathrm{CE},t},
        \\[3pt]
        \left.
        \nabla_{\theta}
        \left(
          \mathcal{L}_{\mathrm{CE},t}
          +\lambda_m\mathcal{L}_{\mathrm{ENLL},t}
        \right)
        \right|_{\widehat e_t=\mathcal{L}_{\mathrm{CE},t}}
        &={}
        \left(
          1+
          \frac{\lambda_m}
               {\mathcal{L}_{\mathrm{CE},t}}
        \right)
        \nabla_{\theta}\mathcal{L}_{\mathrm{CE},t}.
    \end{aligned}
\end{equation}
These identities require pointwise equality to the realized loss, not merely
the context-averaged reference optimum in
Equation~\ref{eq:token_error_population_optimum}. At pointwise equality, IS
adds no gradient to CE for the current error-aware routed model; this does
not imply equality to the native-router baseline update. ENLL still amplifies
that model's CE gradient by the displayed inverse-loss factor. The factor
alone does not determine absolute gradient magnitudes across tokens.
For multiple supervised layers, the same conclusion holds if every layer's
prediction matches the token loss; otherwise the layer contributions must
be averaged as in Equation~\ref{eq:averaged_training_losses}.

Away from equality, the IS coefficient on the CE path is
$1+\lambda_m(1/\widehat e_t-1/\mathcal L_{\mathrm{CE},t})$, which can be
negative; the prediction-gradient term must also be included. Thus, the
analysis does not prove that every joint update decreases CE. Below the IS
floor the IS gradient through the observed loss is zero, while its
prediction gradient uses $\mathcal L^+_{\mathrm{CE},t}$; the floor boundary
requires a subgradient convention. Finally, along the idealized path
$\widehat e_t=\mathcal L_{\mathrm{CE},t}\to0$, ENLL equals
$1+\log\mathcal L_{\mathrm{CE},t}$ and is unbounded below. This is a
property of the unconstrained joint objective, not evidence that a finite
training run attains that limit.

\subsection{Started-Log Routing Geometry}
\label{app:shifted_log_geometry}

This section derives the probability-space form of the attenuation mechanism
and characterizes how predicted error changes expert ranking. From
Equation~\ref{eq:attenuated_routing_score}, exponentiating the adjusted logit
gives
\begin{equation}\label{eq:appendix_shifted_log_exponential}
    \exp(\widetilde{a}_{t,i})
    =
    \exp(a_{t,i})
    \left(1+\frac{\widehat e_{t,i}}{\tau}\right)^{-\gamma}.
\end{equation}
Using the native affinity definition in Equation~\ref{eq:native_affinity} and
cancelling its common softmax normalizer yields
\begin{align}
    \widetilde p_{t,i}
    & =
    \frac{\exp(\widetilde{a}_{t,i})}
         {\sum_{j=1}^{N}\exp(\widetilde{a}_{t,j})}
    \nonumber\\
    & =
    \frac{
      \exp(a_{t,i})
      \left(1+\widehat e_{t,i}/\tau\right)^{-\gamma}
    }{
      \sum_{j=1}^{N}
      \exp(a_{t,j})
      \left(1+\widehat e_{t,j}/\tau\right)^{-\gamma}
    }
    \nonumber\\
    & =
    \frac{
      p_{t,i}
      \left(1+\widehat e_{t,i}/\tau\right)^{-\gamma}
    }{
      \sum_{j=1}^{N}
      p_{t,j}
      \left(1+\widehat e_{t,j}/\tau\right)^{-\gamma}
    }.
    \label{eq:appendix_shifted_log_derivation}
\end{align}
This establishes Equation~\ref{eq:attenuated_probability}.

For two experts, the common normalizer cancels from their routing odds:
\begin{equation}\label{eq:appendix_shifted_log_pairwise}
    \frac{\widetilde p_{t,i}}{\widetilde p_{t,j}}
    =
    \frac{p_{t,i}}{p_{t,j}}
    \left(
      \frac{\tau+\widehat e_{t,i}}
           {\tau+\widehat e_{t,j}}
    \right)^{-\gamma}.
\end{equation}
For $\gamma>0$, if $\widehat e_{t,i}>\widehat e_{t,j}$, attenuation reduces the routing odds
of expert $i$ relative to expert $j$. If their predicted errors are equal, the
common factor cancels and their native affinity ratio is preserved. The method
therefore changes expert ranking through relative differences in predicted
error rather than through a uniform shift shared by all experts. Individual
normalized probabilities need not all decrease: their changes also depend
on the common normalizer. At $\gamma=0$, native probabilities are recovered.

The reference scale $\tau$ makes the transformation sensitive to the magnitude
of predicted error relative to a fixed operating scale. For any multiplicative
rescaling $\xi>0$,
\begin{equation}\label{eq:appendix_shifted_log_rescaling}
    \phi_{\tau}(\xi\widehat e_{t,i})
    =
    \phi_{\tau/\xi}(\widehat e_{t,i}).
\end{equation}
Thus, rescaling all error predictions while holding $\tau$ fixed can change the
route. This differs from pure logarithmic attenuation, for which a common
multiplicative rescaling contributes only a shared logit shift that cancels
under softmax.

The local sensitivities with respect to predicted error and log-error are
\begin{equation}\label{eq:appendix_shifted_log_derivatives}
    \frac{\partial \widetilde{a}_{t,i}}
         {\partial\widehat e_{t,i}}
    =
    -\frac{\gamma}{\tau+\widehat e_{t,i}},
    \qquad
    \frac{\partial \widetilde{a}_{t,i}}
         {\partial\log\widehat e_{t,i}}
    =
    -\gamma
    \frac{\widehat e_{t,i}}{\tau+\widehat e_{t,i}}.
\end{equation}
Sensitivity to predicted error is therefore bounded by $\gamma/\tau$ near
zero, while sensitivity in log-error coordinates increases smoothly toward
$\gamma$.

The two asymptotic regimes make the role of $\tau$ explicit:
\begin{equation}\label{eq:appendix_shifted_log_limits}
    \phi_{\tau}(\widehat e_{t,i})
    =
    \begin{cases}
      \displaystyle \frac{\widehat e_{t,i}}{\tau}
      +\mathcal{O}\!\left(\frac{\widehat e_{t,i}^2}{\tau^2}\right),
      & \widehat e_{t,i}/\tau\to0,\\[8pt]
      \displaystyle \log\frac{\widehat e_{t,i}}{\tau}
      +\mathcal{O}\!\left(\frac{\tau}{\widehat e_{t,i}}\right),
      & \widehat e_{t,i}/\tau\to\infty.
    \end{cases}
\end{equation}
At low error, attenuation is approximately linear and bounded in sensitivity.
At high error, it recovers logarithmic relative-error routing up to a common
shift.

The reference scale $\tau$ and attenuation strength $\gamma$ have different roles:
$\tau$ sets the transition between the linear and logarithmic regimes, whereas
$\gamma$ scales the overall routing adjustment. We set both to one in the
default configuration. This routing transformation does not change the direct
IS or ENLL scalar prediction optimum for fixed observed losses derived in
Appendix~\ref{app:token_error_optimality}; it changes how the learned prediction
affects expert selection and can therefore change joint training dynamics.

\subsection{Error-to-Routing Transformation Ablation}
\label{app:error_routing_transform_analysis}

The started-log route in Equation~\ref{eq:attenuated_routing_score} uses the
absolute operating range of the error head relative to $\tau$. We compare it
with the scale-invariant pure-log control
\begin{equation}\label{eq:shifted_log_ablation_score}
    \widetilde{a}_{t,i}^{\log}
    =
    a_{t,i}
    -\gamma\log\widehat e_{t,i}.
\end{equation}
The pure logarithm applies the same routing adjustment to a fixed error ratio
at every absolute error scale. The started logarithm instead contracts the
low-error region and removes the singular routing sensitivity near zero.
Appendix~\ref{app:shifted_log_geometry} derives its pairwise form, probability
interpretation, and gradients.

Table~\ref{tab:error_routing_transform_ablation} compares the two transforms on
ARC-Challenge with Granite, IS supervision, $\gamma=1$, error-head learning
rate $10^{-5}$, and the fixed epoch-3 checkpoint. Correlations are calculated
on the test population after averaging each layer--token-identity group over seeds
42--44.

\begin{table}[H]
    \setlength{\belowcaptionskip}{\abovecaptionskip}
    \setlength{\abovecaptionskip}{0pt}
    \caption{Error-to-routing transformation ablation on ARC-Challenge with
    Granite and IS supervision. Accuracy, NLL, and pointwise Pearson $r$ are
    epoch-3 means over seeds 42--44. Group-level Pearson $r$ and Spearman
    $\rho$ are computed after averaging each layer--token-identity group across the same
    seeds. Bold marks the better observed value in each column and does not
    imply statistical significance.}
    \label{tab:error_routing_transform_ablation}
    \centering
    \small
    \setlength{\tabcolsep}{5.5pt}
    \renewcommand{\arraystretch}{1.12}
    \begin{tabular}{@{}lrrrrr@{}}
        \toprule
        Routing transform
        & Pointwise $r$
        & Group $r$
        & Group $\rho$
        & Test NLL
        & Accuracy (\%) \\
        \midrule
        $\log\widehat e$
        & 0.1586 & 0.3975 & 0.4977 & 0.9723 & \textbf{67.27} \\
        $\log(1+\widehat e)$
        & \textbf{0.2486} & \textbf{0.4521} & \textbf{0.5637}
        & \textbf{0.9673} & 66.87 \\
        \bottomrule
    \end{tabular}
\end{table}

\begin{figure}[H]
    \centering
    \includegraphics[width=\linewidth]{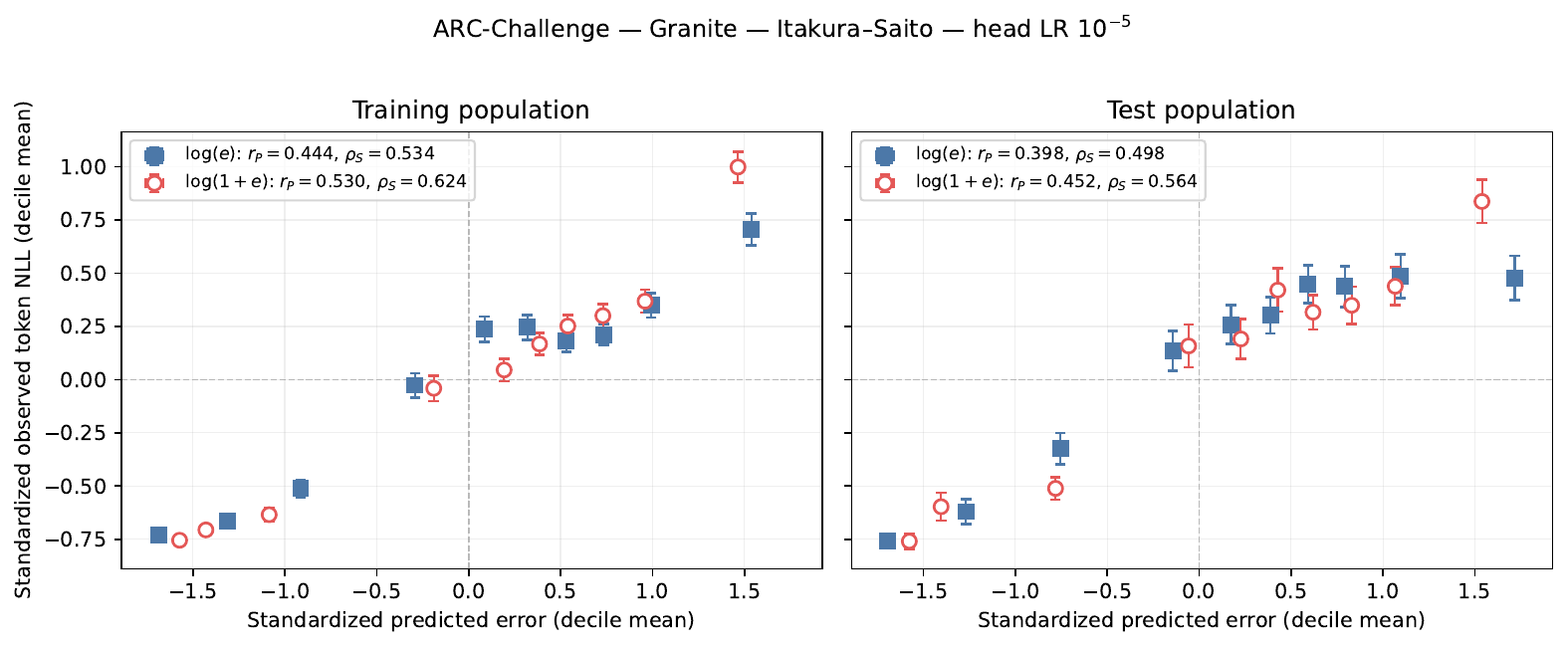}
    \caption{Standardized aggregate error prediction for the pure-log control
    and the started-log route on ARC-Challenge with Granite, IS supervision,
    error-head learning rate $10^{-5}$, and the fixed epoch-3 checkpoint.
    Points are equal-count decile means of seed-averaged layer--token-identity
    groups; whiskers show standard errors across groups within each decile.
    Legend values report Pearson $r_P$ and Spearman $\rho_S$ over all groups,
    not over the ten displayed points.
    Training-population standardization is applied unchanged to test.}
    \label{fig:error_routing_transform_standardized}
\end{figure}

Figure~\ref{fig:error_routing_transform_standardized} visualizes the same
comparison. For each transform, we average every layer--token-identity group over
seeds 42--44 and form equal-count deciles ordered by predicted
error. Predicted and observed group means are standardized using the
corresponding training-population moments, which are then applied unchanged to
the test population. This removes the large difference in raw prediction scale
while preserving each transform's ordering and out-of-sample shift.

Started-log routing increases group-level test Pearson correlation from $0.398$ to
$0.452$ and Spearman correlation from $0.498$ to $0.564$. The corresponding
training correlations increase from $0.444$ to $0.530$ and from $0.534$ to
$0.624$, respectively. It also improves pointwise correlation and test NLL,
while the pure-log control has $0.40$ percentage points higher answer-choice
accuracy. The comparison therefore supports a conditioning interpretation:
the started-log transform permits a wider raw prediction range while
compressing its contribution to routing, and the resulting head ranks token
difficulty more consistently. The different accuracy ordering shows that error
estimation and useful expert selection remain distinct requirements.

The compared checkpoints also use different neutral error-head
initializations: the pure-log route starts at $\widehat e=1$, whereas the
started-log route uses the zero-parameter softplus initialization. The observed
contrast therefore combines transformation and initialization effects and
should not be interpreted as a fully isolated causal estimate.

\subsection{TES Hyperparameter Sensitivity}
\label{app:tes_hyperparameter_sensitivity}

TES introduces two method-specific optimization choices: the learning rate of
the token-error head and the attenuation scale $\gamma$. We evaluate their
sensitivity on Granite ARC-Challenge using started-log IS supervision while
keeping the backbone, data, training schedule, and seed cohort fixed.

\input{tables/tes_hyperparameter_sensitivity}

Table~\ref{tab:tes_hyperparameter_sensitivity} shows that the selected
defaults---error-head learning rate $10^{-5}$ and $\gamma=1$---give the highest
observed accuracy in the corresponding sweeps. Nearby settings
produce similar results, while the $\gamma$ sweep also shows that minimizing
choice NLL and maximizing answer-choice accuracy need not select the same value.
These experiments cover IS on Granite ARC-Challenge.

\subsection{Experimental Configuration}
\label{app:experimental_configuration}

Table~\ref{tab:experimental_configuration} reports the complete configuration
used for the primary downstream experiments. Within each backbone and dataset,
all methods use the same registered splits, prompt, candidate scoring rule,
adaptation surface, and evaluation procedure.

The registered manifests define custom partitions rather than subsamples of
official test sets. The split builder pools the declared source partitions,
filters invalid examples, and assigns examples deterministically with seed 42.
Source identifiers are retained, with disjoint IDs across the training,
validation, and test manifests. The released split metadata record the source
partitions and their allocation; examples from an official test partition can
therefore occur in the custom training partition. These results should not be
interpreted as official-test evaluations.

Our evaluator uses the \texttt{manifest\_answer\_text\_choice\_ce} profile.
The prompt is \texttt{Question: <question>} followed by a newline and
\texttt{Answer:}; each candidate appends one space followed by its answer
text, without a chat wrapper or in-context demonstrations. Candidates are
ranked by mean conditional token log-probability. Accuracy is the fraction of
correctly selected answers, denoted \texttt{acc\_norm} in the evaluation
artifacts. The primary adaptation objective is CE over these answer-choice
scores, whereas TES and ACS use observed next-token CE as their auxiliary
target. Evaluation uses the repository's manifest-based candidate scorer
rather than an unmodified benchmark-harness invocation.

\begin{table}[t]
    \setlength{\belowcaptionskip}{\abovecaptionskip}
    \setlength{\abovecaptionskip}{0pt}
    \caption{Backbone, optimization, and method configurations for the primary
    downstream evaluation. ARC denotes ARC-Challenge; ``others'' denotes
    OpenBookQA, SciQ, and MedMCQA.}
    \label{tab:experimental_configuration}
    \centering
    \small
    \setlength{\tabcolsep}{2.5pt}
    \renewcommand{\arraystretch}{1.12}
    \begin{tabular}{@{}p{0.29\linewidth}p{0.32\linewidth}p{0.32\linewidth}@{}}
        \toprule
        Setting & Granite 3.1 3B-A800M & OLMoE-1B-7B-SFT \\
        \midrule
        Training examples
            & 2,000 (ARC); 4,992 (others)
            & 2,000 (ARC); 5,000 (others) \\
        Validation / test examples & 50 / 500 & 50 / 500 \\
        LoRA rank / $\alpha$ / dropout & 8 / 8 / 0.05 & 8 / 8 / 0.05 \\
        LoRA targets
            & Q/K/V and routed-expert projections
            & Q/K/V/O and routed-expert projections \\
        LoRA learning rate & $10^{-4}$ & $2\times10^{-5}$ \\
        Batch / accumulation & 8 / 2 & 8 / 1 \\
        Optimizer & AdamW & AdamW \\
        Schedule / warmup & cosine / 0.05 & linear / 0.03 \\
        Weight decay / gradient clip & 0 / 1 & 0 / 1 \\
        Precision & BF16 & BF16 \\
        IS observation floor $\varepsilon_{\mathrm{IS}}$ & $10^{-8}$ & $10^{-8}$ \\
        Steps per epoch
            & 125 (ARC); 312 (others)
            & 250 (ARC); 625 (others) \\
        Epochs / seeds & 3 / 42, 43, 45 & 3 / 42, 43, 45 \\
        Supervised MoE scope & final sparse layer & final sparse layer \\
        Native router / auxiliary loss & frozen / 0 & frozen / 0 \\
        \midrule
        Method & Objective and method-specific configuration & Evaluation route \\
        \midrule
        CE & task CE only & native affinity \\
        Dual Affinity
            & task CE; copied-head LR $10^{-3}$; mixing weight 0.5
            & Dual Affinity \\
        TES--IS
            & task CE $+\,10^{-3}\mathcal L_{\mathrm{IS}}$;
              error-head LR $10^{-5}$
            & started-log attenuation, $\gamma=\tau=1$ \\
        TES--ENLL
            & task CE $+\,10^{-3}\mathcal L_{\mathrm{ENLL}}$;
              error-head LR $10^{-5}$
            & started-log attenuation, $\gamma=\tau=1$ \\
        ACS--IS
            & task CE $+\,10^{-3}\mathcal L_{\mathrm{ACS\text{-}IS}}$; no method head
            & native affinity \\
        ACS--ENLL
            & task CE $+\,10^{-3}\mathcal L_{\mathrm{ACS\text{-}ENLL}}$; no method head
            & native affinity \\
        \bottomrule
    \end{tabular}
\end{table}

Every epoch checkpoint is saved. Epoch 3 is the predeclared primary endpoint.
For the secondary robustness analysis, one checkpoint is selected independently
for each seed by minimum length-normalized answer-choice NLL on the 50-example
validation split; exact ties select the earlier epoch. The selection is fixed
before test evaluation, and test metrics do not enter the selection rule.

\subsection{Cross-Dataset Error-Prediction Analysis}
\label{app:cross_dataset_error_prediction}

\begin{figure}[t]
    \centering
    \begin{minipage}[t]{0.495\linewidth}
        \centering
        \includegraphics[width=\linewidth]{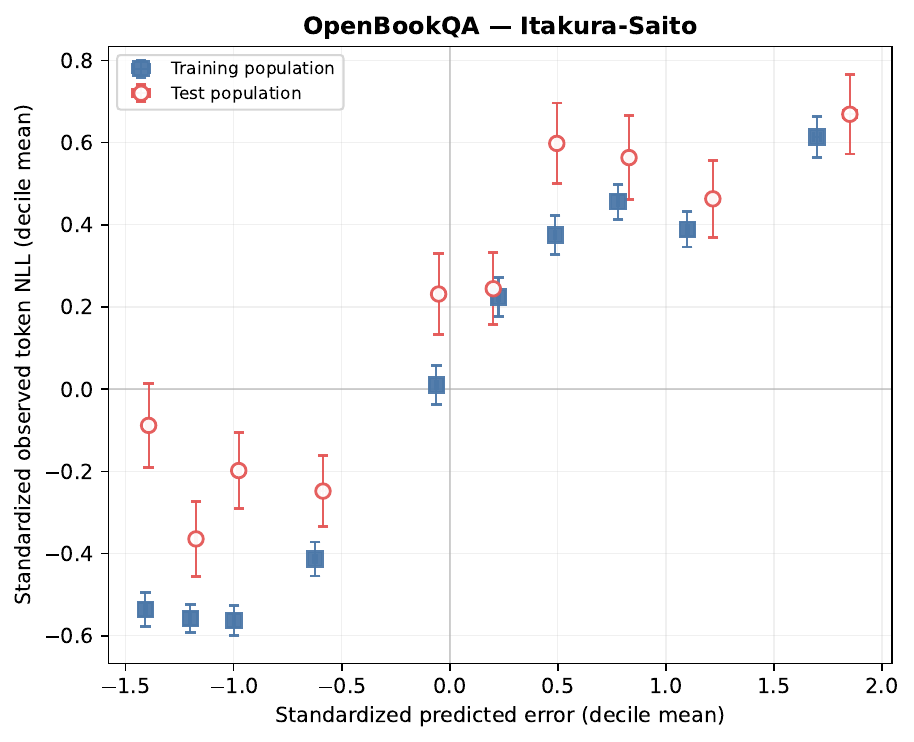}
        \vspace{-0.8em}
        \textbf{(a) Itakura--Saito}
    \end{minipage}\hfill
    \begin{minipage}[t]{0.495\linewidth}
        \centering
        \includegraphics[width=\linewidth]{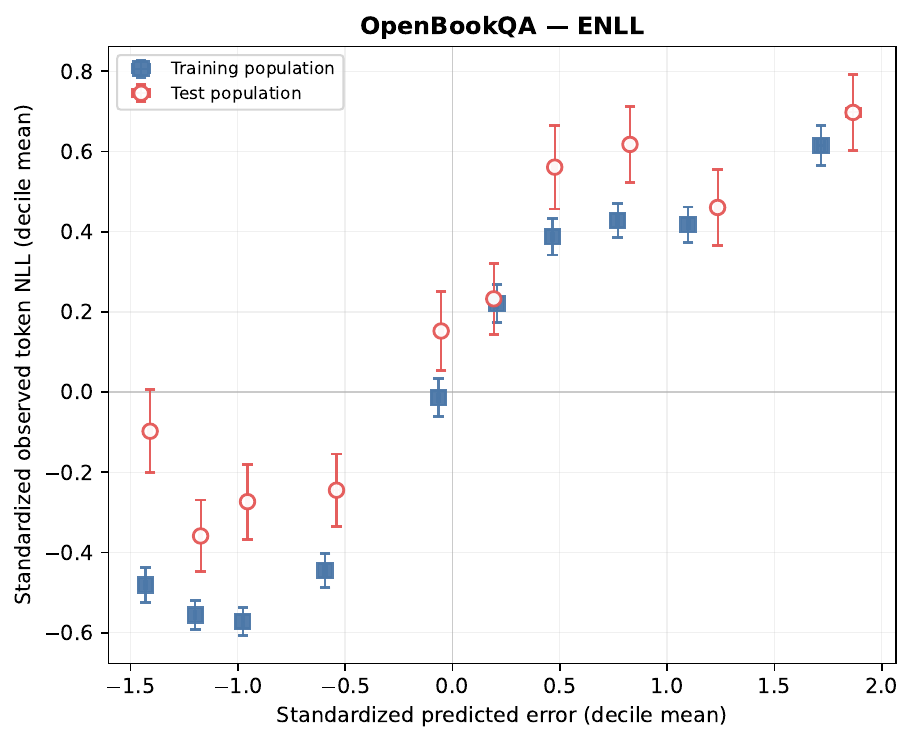}
        \vspace{-0.8em}
        \textbf{(b) ENLL}
    \end{minipage}
    \caption{Standardized aggregate error prediction on OpenBookQA for the
    Granite \(\log(1+\widehat e)\) runs. Points are equal-count decile
    means after averaging layer--token-identity groups over seeds 42--44; whiskers
    show \(\pm1\) standard error across groups. Blue squares and red circles denote the
    training and test populations. Axis limits are selected automatically for
    each panel.}
    \label{fig:openbookqa_error_prediction}
\end{figure}

\begin{figure}[t]
    \centering
    \begin{minipage}[t]{0.495\linewidth}
        \centering
        \includegraphics[width=\linewidth]{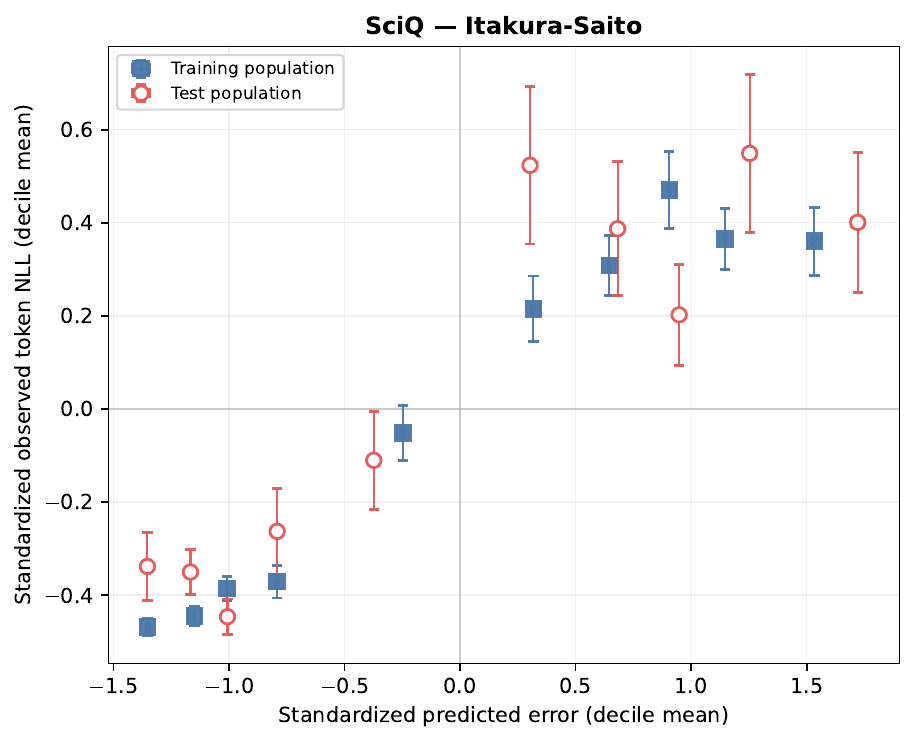}
        \vspace{-0.8em}
        \textbf{(a) Itakura--Saito}
    \end{minipage}\hfill
    \begin{minipage}[t]{0.495\linewidth}
        \centering
        \includegraphics[width=\linewidth]{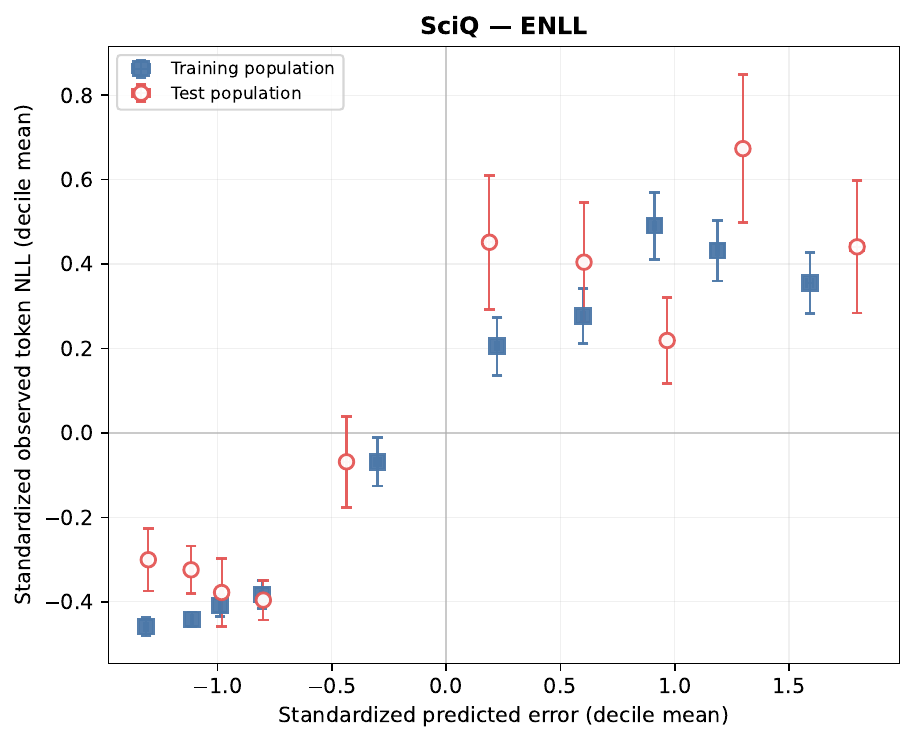}
        \vspace{-0.8em}
        \textbf{(b) ENLL}
    \end{minipage}
    \caption{Standardized aggregate error prediction on SciQ for the
    Granite \(\log(1+\widehat e)\) runs. Plot construction and visual encoding
    follow Figure~\ref{fig:openbookqa_error_prediction}.}
    \label{fig:sciq_error_prediction}
\end{figure}

\begin{figure}[t]
    \centering
    \begin{minipage}[t]{0.495\linewidth}
        \centering
        \includegraphics[width=\linewidth]{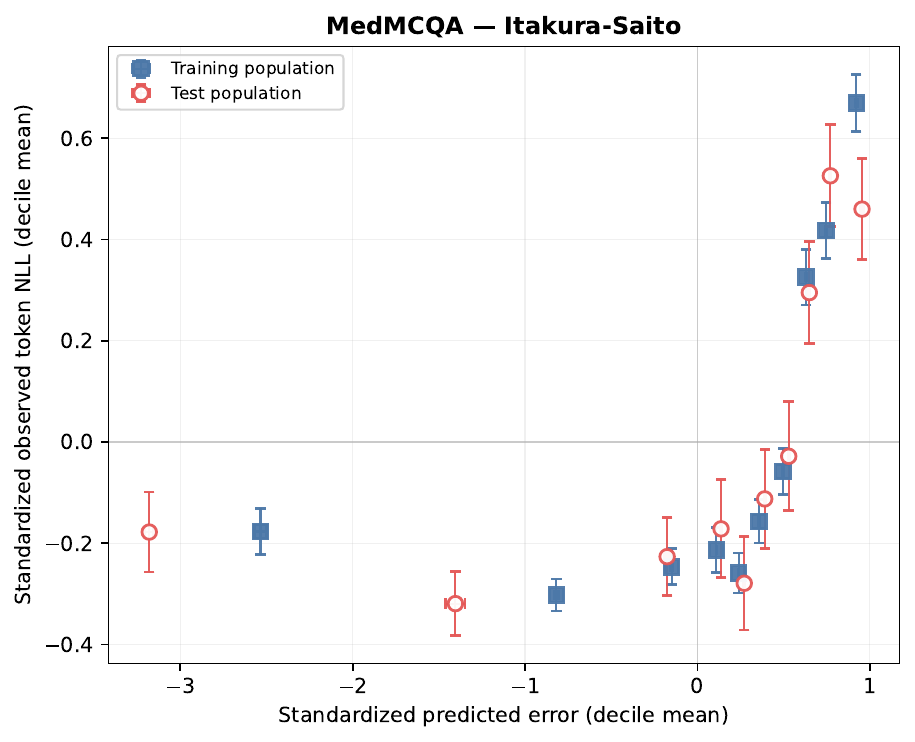}
        \vspace{-0.8em}
        \textbf{(a) Itakura--Saito}
    \end{minipage}\hfill
    \begin{minipage}[t]{0.495\linewidth}
        \centering
        \includegraphics[width=\linewidth]{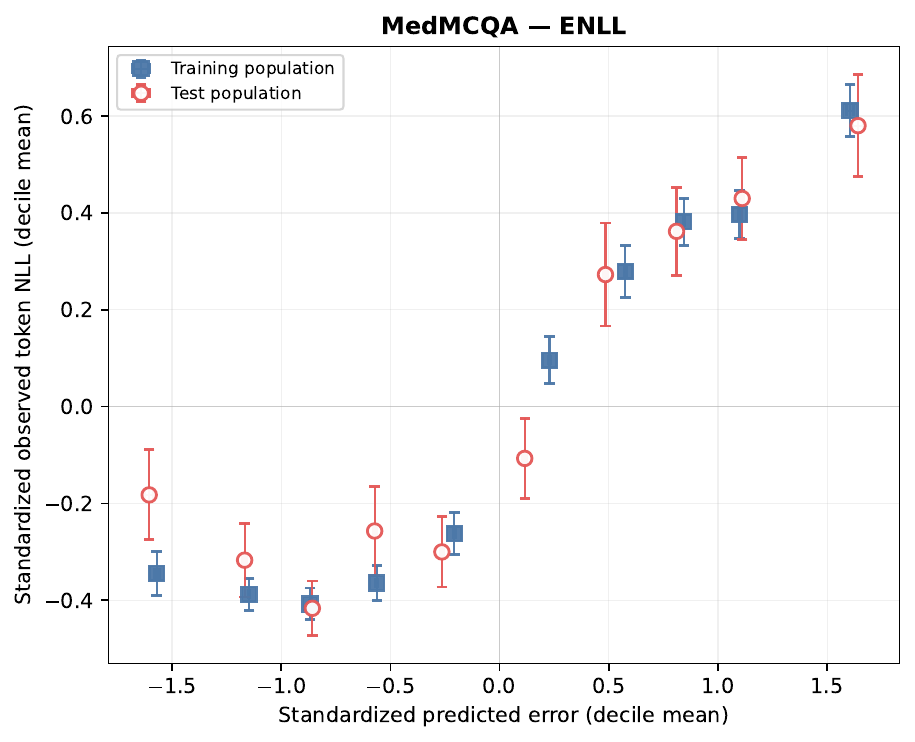}
        \vspace{-0.8em}
        \textbf{(b) ENLL}
    \end{minipage}
    \caption{Standardized aggregate error prediction on MedMCQA for the
    Granite \(\log(1+\widehat e)\) runs. Plot construction and visual encoding
    follow Figure~\ref{fig:openbookqa_error_prediction}.}
    \label{fig:medmcqa_error_prediction}
\end{figure}

Figures~\ref{fig:openbookqa_error_prediction}--
\ref{fig:medmcqa_error_prediction} extend the ARC-Challenge analysis in
Figure~\ref{fig:arc_error_prediction} to the remaining datasets using the
Granite \(\log(1+\widehat e)\) experiments. For each objective and
dataset, we use the epoch-3 checkpoint with \(\gamma=1\), error-head learning
rate \(10^{-5}\). Within each seed, we average
predicted and observed errors over the contextual occurrences in each
layer--token-identity group. We then average group means across seeds and form
equal-count deciles ordered by predicted error. Predicted and observed group
means are standardized by their respective
training-population moments, which are then applied unchanged to the test
population. Each panel uses automatically selected axis limits for visibility.
The analysis evaluates the aggregate prediction; it does not interpret the
expert-indexed scores as observed counterfactual expert errors.

Both OpenBookQA objectives retain positive out-of-sample ordering. Across test
groups, Pearson/Spearman correlations are \(0.316/0.333\) for IS and
\(0.321/0.344\) for ENLL. The highest predicted-error decile also has a larger
standardized observed NLL than the lowest decile for both objectives. Because
the axes are standardized separately from the training reference, this figure
assesses association and distribution shift rather than absolute NLL
calibration.

SciQ also shows positive test-set group association, with Pearson/Spearman
correlations of \(0.300/0.378\) for IS and \(0.308/0.377\) for ENLL. The decile
trajectories are not monotone at every adjacent transition, but both objectives
separate their lowest and highest predicted-error deciles in the expected
direction.

MedMCQA ENLL retains positive test-set group association
(Pearson/Spearman \(0.297/0.353\)) and a positive endpoint separation. The IS
aggregate is weaker (\(0.142/0.240\)). Across the evaluated datasets, the normalized
plots show that both objectives learn a positive token-error ordering, while
the strength of that ordering remains dataset dependent.

\subsection{Affinity-Concentration Supervision}
\label{app:affinity_softmax_error_supervision}

Affinity-concentration supervision (ACS) is a headless alternative to the
learned error-head method. It uses token-loss supervision to shape the native
router distribution directly. ACS adds no
error-specific parameters and applies no error-dependent transformation at
inference. Expert selection and aggregation therefore continue to follow the
native route in Equations~\ref{eq:native_affinity}--
\ref{eq:native_moe_output}.

For supervised token position $t$ in layer $l$, let $\mathcal S_t^{(l)}$ be the native top-$K$
expert set. We normalize the native affinity probabilities within this set and
measure their squared concentration:
\begin{equation}\label{eq:appendix_affinity_concentration}
    \bar p_{t,i}^{(l)}
    =
    \frac{p_{t,i}^{(l)}}
         {\sum_{j\in\mathcal S_t^{(l)}}p_{t,j}^{(l)}},
    \qquad
    C_t^{(l)}
    =
    \sum_{i\in\mathcal S_t^{(l)}}
    \left(\bar p_{t,i}^{(l)}\right)^2.
\end{equation}
The concentration satisfies $1/K\leq C_t^{(l)}\leq1$: Cauchy--Schwarz gives
$1=(\sum_i\bar p_{t,i}^{(l)})^2\leq K C_t^{(l)}$, and
$\sum_i(\bar p_{t,i}^{(l)})^2\leq(\sum_i\bar p_{t,i}^{(l)})^2=1$,
where both sums are over the selected set. The lower endpoint corresponds
to equal selected affinities. For $K>1$ and finite softmax logits, all selected
probabilities are positive, so $C_t^{(l)}<1$; the upper endpoint is approached
as affinity concentrates on one expert. For $K=1$, concentration is identically
one and supplies no direct concentration gradient. Equivalently,
$1/C_t^{(l)}$ is the effective number of selected experts and
$-\log C_t^{(l)}$ is their order-two R\'enyi entropy
\citep{renyi1961measures,hill1973diversity}. Thus, $C_t^{(l)}$ is a bounded description of
router sharpness, not a calibrated estimate of token or expert error.

For evaluation, we compute Equation~\ref{eq:appendix_affinity_concentration}
at every valid supervised token position using the exact expert set executed
by the native forward pass. Let $\mathcal T_{\mathcal D}$ denote these token
positions for evaluation split $\mathcal D$. We average over these positions
within each checkpoint, then report the mean and sample standard deviation
across seeds, weighting each seed equally. Appendix~
\ref{app:acs_concentration_diagnostics} defines the reported statistics.
When multiple layers are supervised, concentration is reported separately
by layer unless a cross-layer average is explicitly identified.
Within-token affinity concentration is distinct from expert-load
concentration across a dataset.

ACS applies the same IS or ENLL form used by the error-head method, but
substitutes the native concentration for the learned aggregate prediction:
\begin{align}
    \mathcal L_{\mathrm{ACS\text{-}IS},t}^{(l)}
    & =
    \frac{\mathcal L^{+}_{\mathrm{CE},t}}{C_t^{(l)}}
    -\log\!\left(
      \frac{\mathcal L^{+}_{\mathrm{CE},t}}{C_t^{(l)}}
    \right)-1,
    \label{eq:appendix_affinity_softmax_is}\\
    \mathcal L_{\mathrm{ACS\text{-}ENLL},t}^{(l)}
    & =
    \log C_t^{(l)}
    +\frac{\mathcal L_{\mathrm{CE},t}}{C_t^{(l)}}.
    \label{eq:appendix_affinity_softmax_enll}
\end{align}
The two objectives have the direct concentration gradients
\begin{align}
    \frac{\partial\mathcal L_{\mathrm{ACS\text{-}IS},t}^{(l)}}
         {\partial C_t^{(l)}}
    &=
    \frac{C_t^{(l)}-\mathcal L^{+}_{\mathrm{CE},t}}
         {\left(C_t^{(l)}\right)^2},
    \label{eq:appendix_affinity_concentration_is_gradient}\\
    \frac{\partial\mathcal L_{\mathrm{ACS\text{-}ENLL},t}^{(l)}}
         {\partial C_t^{(l)}}
    &=
    \frac{C_t^{(l)}-\mathcal L_{\mathrm{CE},t}}
         {\left(C_t^{(l)}\right)^2}.
    \label{eq:appendix_affinity_concentration_enll_gradient}
\end{align}
When the IS floor is inactive, these gradients coincide. With the selected
set fixed, its normalization cancels the all-expert softmax denominator:
$\bar p_{t,i}^{(l)}=\exp(a_{t,i}^{(l)})/
\sum_{j\in\mathcal S_t^{(l)}}\exp(a_{t,j}^{(l)})$.
Consequently, differentiating the sum of squared selected probabilities gives
\begin{equation}\label{eq:appendix_affinity_concentration_logit_gradient}
    \frac{\partial C_t^{(l)}}{\partial a_{t,i}^{(l)}}
    =
    2\bar p_{t,i}^{(l)}
    \left(\bar p_{t,i}^{(l)}-C_t^{(l)}\right),
    \qquad i\in\mathcal S_t^{(l)}.
\end{equation}
For $i\notin\mathcal S_t^{(l)}$, the local derivative is zero while the
selected set remains fixed; top-$K$ boundaries are nondifferentiable.
Multiplying this derivative by the corresponding concentration derivative
above gives the direct auxiliary gradient for a selected logit, holding CE
fixed. For this path alone, gradient descent favors greater concentration
when the target exceeds $C_t^{(l)}$, and less concentration when it is smaller.
At exactly uniform selected affinity, however,
$\bar p_{t,i}^{(l)}=C_t^{(l)}=1/K$ and every concentration-logit derivative
is zero. Near this point, and near single-expert saturation, the direct
gradient can be small. Thus, the scalar gradient's sign does not guarantee
a concentration change in a parameter update, which also includes CE and
shared-parameter effects. ACS does not identify which unselected expert
would reduce the loss and supplies no counterfactual expert-error target.

The bounded range also prevents ACS from inheriting the unrestricted
error-prediction interpretation of
Section~\ref{subsec:token_error_supervision}. Across the contextual occurrences
$c$ of a fixed target token identity $t'$, consider one scalar concentration
reference $C_{t'}^{(l)}$ held constant inside the average, as in
Appendix~\ref{app:token_error_optimality}. Its expected-objective derivative
is $(C_{t'}^{(l)}-\mathbb E_{c\mid t=t'}[\mathcal L^+_{\mathrm{CE},t,c}])/
(C_{t'}^{(l)})^2$ for IS, with the unfloored loss for ENLL. Its sign changes
at the mean target, so minimization over the closed interval $[1/K,1]$ gives
\begin{equation}\label{eq:appendix_affinity_concentration_optimum}
    C_{\mathrm{IS},t'}^{(l)\star}
    =
    \operatorname{clip}\!\left(
      \mathbb E_{c\mid t=t'}[\mathcal L^{+}_{\mathrm{CE},t,c}],
      \frac{1}{K},1
    \right),
    \qquad
    C_{\mathrm{ENLL},t'}^{(l)\star}
    =
    \operatorname{clip}\!\left(
      \mathbb E_{c\mid t=t'}[\mathcal L_{\mathrm{CE},t,c}],
      \frac{1}{K},1
    \right).
\end{equation}
Here $\operatorname{clip}$ restricts its first argument to $[1/K,1]$.
These scalar reference optima are not the average of independently optimized
contextual concentrations: averaging and clipping do not generally commute.
The native router further couples occurrences through shared parameters.
For $K>1$, when the scalar reference optimum is one, softmax routing
approaches the objective's infimum as selected logit differences grow;
no finite logits attain it. Expected losses at least one
therefore share the same upper-boundary reference.
Moreover, because the observed token loss is
not detached, IS and ENLL retain different gradients through the language
model even when their direct concentration gradients agree. For
$\mathcal L_{\mathrm{CE},t}>\varepsilon_{\mathrm{IS}}$,
\begin{equation}\label{eq:appendix_affinity_observation_gradients}
    \frac{\partial\mathcal L_{\mathrm{ACS\text{-}IS},t}^{(l)}}
         {\partial\mathcal L_{\mathrm{CE},t}}
    =
    \frac{1}{C_t^{(l)}}-
    \frac{1}{\mathcal L_{\mathrm{CE},t}},
    \qquad
    \frac{\partial\mathcal L_{\mathrm{ACS\text{-}ENLL},t}^{(l)}}
         {\partial\mathcal L_{\mathrm{CE},t}}
    =
    \frac{1}{C_t^{(l)}}.
\end{equation}
Below the IS floor, the IS derivative through the observed loss is zero;
the ENLL expression is unchanged. The full chain rule adds this CE path
to the concentration path, exactly as in
Appendix~\ref{app:attached_token_loss_gradients} with
$\widehat e_t$ replaced by $C_t^{(l)}$.

The ACS training objective follows Equation~\ref{eq:total_loss}, with
$m\in\{\mathrm{ACS\text{-}IS},\mathrm{ACS\text{-}ENLL}\}$ and the generic
coefficient $\lambda_m$. Through trainable parameters that affect the affinity
logits, ACS can alter future native routes, but it does not replace the native
routing algorithm during either training or evaluation.

Table~\ref{tab:downstream_accuracy} reports the complete Granite and OLMoE ACS
results. The two ACS objectives remain close across datasets, and neither is
uniformly preferred. On Granite, ACS-IS attains the highest observed mean on
ARC-Challenge and SciQ; the corresponding OLMoE results are competitive but do
not exceed learned-head IS on ARC-Challenge or OpenBookQA.

For the frozen-affinity ARC-Challenge epoch-3 runs with final-layer ACS and
$\lambda_m=10^{-3}$, mean concentration remains
between $0.1394$ and $0.1409$, close to the uniform-routing endpoint
$1/K=0.125$ for $K=8$, while its Pearson correlation with observed token CE is
weak and negative ($-0.04$ to $-0.14$). These diagnostics do not establish
concentration as a token-error predictor or isolate whether the accuracy gains
arise from concentration regularization, attached-loss gradient shaping, or both.

\subsection{ACS Supervision-Coefficient Sensitivity}
\label{app:acs_coefficient_sensitivity}

The primary comparison fixes $\lambda_m=10^{-3}$ across datasets. We test the
sensitivity of ACS to this choice by comparing it with $\lambda_m=10^{-2}$
under otherwise matched final-layer configurations. We retain every evaluated
dataset in Table~\ref{tab:acs_coefficient_sensitivity}, including those for
which the larger coefficient reduces accuracy.

\input{tables/acs_coefficient_sensitivity}

The larger coefficient improves five of eight Granite trainable-affinity
objective--dataset cells and four of eight cells in each OLMoE affinity
setting. The largest increases are $1.67$ percentage points for Granite
ACS--ENLL on OpenBookQA and $1.34$ points for frozen-affinity OLMoE ACS--ENLL
on SciQ. Table~\ref{tab:olmoe_acs_coefficient_screen} also shows that increasing
the coefficient further to $10^{-1}$ sharply reduces OLMoE ARC-Challenge
accuracy. Thus, $10^{-2}$ is a useful dataset-dependent alternative rather
than a uniformly better default.

The OLMoE trainable-affinity comparison uses validation-NLL-selected checkpoints.

\subsection{Affinity Supervision Depth Ablation}
\label{app:affinity_router_layer_ablation}

Section~\ref{subsec:supervision_depth} isolates supervision depth while keeping
native affinity frozen. Here, we study the coupled setting in which ACS
supervision and native-affinity training are applied to the same layer scope.
We compare the final layer, first half, last half, and full MoE stack.
Consequently, differences between scopes reflect both the placement of
token-loss supervision and the number of affinity routers being optimized.

\input{tables/affinity_router_layer_ablation}

Table~\ref{tab:affinity_router_layer_ablation} reports the fixed epoch-3
Granite endpoint at $\lambda_m=10^{-3}$. For Granite, final-layer
training gives the highest ACS accuracy for both objectives on all four
datasets, whereas all-layer training is consistently weakest. The matched
Router-CE control also degrades as affinity training broadens, showing that the
all-layer decline is not specific to ACS. Relative to the matched control at
each scope, the clearest additional benefit from ACS occurs on ARC-Challenge
under final-layer and last-half training. These results support narrow affinity
supervision rather than indiscriminate optimization of every router.

The OLMoE panel reports validation-NLL-selected checkpoints at
$\lambda_m=10^{-2}$ together with the matched Router-CE controls.
These results do not identify a universal placement:
ARC-Challenge and SciQ favor narrow or final-layer supervision, whereas
OpenBookQA and MedMCQA favor broader or half-depth scopes depending on the
objective. Relative to Router CE, ACS helps on SciQ, is roughly neutral on
ARC-Challenge and MedMCQA, and hurts on OpenBookQA. We therefore do not compare
this panel directly with the frozen-affinity results in
Section~\ref{subsec:supervision_depth}, which uses fixed epoch-3 checkpoints.

\subsection{Affinity-Concentration Diagnostics}
\label{app:acs_concentration_diagnostics}

Accuracy alone does not establish whether ACS changes the concentration it
directly supervises. Table~\ref{tab:acs_concentration_diagnostics} reports the
mean selected-set concentration $\bar C_{\mathcal D}$, its reciprocal effective
expert count $N_{\mathrm{eff},\mathcal D}$, and the token-level Pearson
correlation $r_{\mathcal D}$
between concentration and realized loss for the final-layer Granite
Router-CE and ACS runs on the ARC-Challenge train and test sets. In this
comparison, the final native affinity router is trainable in all three methods,
and ACS uses $\lambda_m=10^{-2}$. This differs from the frozen-affinity,
$\lambda_m=10^{-3}$ diagnostics in
Appendix~\ref{app:affinity_softmax_error_supervision}.

\begin{samepage}
For one checkpoint and split
$\mathcal D\in\{\mathrm{train},\mathrm{test}\}$, let
$\mathcal T_{\mathcal D}$ denote the evaluated token positions. All quantities below are
computed at the final sparse MoE layer, and we suppress the layer index as in
Section~\ref{subsec:affinity_concentration_supervision}. For position
$t\in\mathcal T_{\mathcal D}$, $\mathcal S_t$ is the native top-$K$ expert set and
$p_{t,i}$ is the affinity probability assigned to expert $i$. The selected-set
probability $\bar p_{t,i}$ and per-token concentration $C_t$ are
\begin{equation}
    \bar p_{t,i}
    =
    \frac{p_{t,i}}
         {\sum_{j\in\mathcal S_t}p_{t,j}},
    \qquad
    C_t
    =
    \sum_{i\in\mathcal S_t}\bar p_{t,i}^{2}.
    \label{eq:diagnostic_token_concentration}
\end{equation}
\end{samepage}
Equation~\ref{eq:diagnostic_token_concentration} is the same order-two
concentration defined in Equation~\ref{eq:affinity_concentration}. Granite uses
$K=8$, so $1/8\leq C_t\leq1$. The checkpoint-level mean concentration and its
reciprocal effective expert count are
\begin{equation}
    \bar C_{\mathcal D}
    =
    \frac{1}{|\mathcal T_{\mathcal D}|}
    \sum_{t\in\mathcal T_{\mathcal D}}C_t,
    \qquad
    N_{\mathrm{eff},\mathcal D}
    =
    \frac{1}{\bar C_{\mathcal D}}.
    \label{eq:diagnostic_split_concentration}
\end{equation}
Thus, $1/K\leq\bar C_{\mathcal D}\leq1$ and
$1\leq N_{\mathrm{eff},\mathcal D}\leq K$: larger $\bar C_{\mathcal D}$ and
smaller $N_{\mathrm{eff},\mathcal D}$ indicate that the selected affinity mass is carried by
fewer experts. We first average concentration over token positions within each
checkpoint as shown in Equation~\ref{eq:diagnostic_split_concentration} and then
take its reciprocal; this differs from averaging $1/C_t$ over positions.

The final column measures the token-level Pearson association between
concentration and realized next-token loss:
\begin{equation}
    r_{\mathcal D}
    =
    \frac{
      \sum_{t\in\mathcal T_{\mathcal D}}
      (C_t-\bar C_{\mathcal D})
      (\mathcal L_{\mathrm{CE},t}-\bar{\mathcal L}_{\mathrm{CE},\mathcal D})
    }{
      \sqrt{\sum_{t\in\mathcal T_{\mathcal D}}(C_t-\bar C_{\mathcal D})^2}
      \sqrt{\sum_{t\in\mathcal T_{\mathcal D}}
      (\mathcal L_{\mathrm{CE},t}-\bar{\mathcal L}_{\mathrm{CE},\mathcal D})^2}
    },
    \qquad
    \bar{\mathcal L}_{\mathrm{CE},\mathcal D}
    =
    \frac{1}{|\mathcal T_{\mathcal D}|}
    \sum_{t\in\mathcal T_{\mathcal D}}\mathcal L_{\mathrm{CE},t}.
    \label{eq:diagnostic_concentration_loss_correlation}
\end{equation}
A positive $r_{\mathcal D}$ means that higher-loss tokens tend to have more concentrated
selected affinity; a negative value indicates the opposite association. For
constant concentration or constant loss, the denominator is zero and the
correlation is undefined, rather than zero. For
each method and split, Table~\ref{tab:acs_concentration_diagnostics} reports
the mean and sample standard deviation of $\bar C_{\mathcal D}$,
$N_{\mathrm{eff},\mathcal D}$, and $r_{\mathcal D}$ across the three training seeds.

\begin{table}[t]
    \setlength{\belowcaptionskip}{\abovecaptionskip}
    \setlength{\abovecaptionskip}{0pt}
    \caption{Final-layer affinity-concentration diagnostics for Granite on
    ARC-Challenge with the native affinity router trained by task CE alone
    (Router CE) or jointly with ACS at $\lambda_m=10^{-2}$.
    Entries are epoch-3 means $\pm$ sample
    standard deviations over seeds 42, 43, and 45. Each seed contains 14,679
    train-token and 3,550 test-token observations.}
    \label{tab:acs_concentration_diagnostics}
    \centering
    \small
    \setlength{\tabcolsep}{2.5pt}
    \renewcommand{\arraystretch}{1.12}
    \begin{tabular*}{\columnwidth}{@{\extracolsep{\fill}}llrrr@{}}
        \toprule
        Method & Split & $\bar C_{\mathcal D}$ & $N_{\mathrm{eff},\mathcal D}$ & $r_{\mathcal D}$ \\
        \midrule
        Router CE & Train & $0.2614\pm0.0279$ & $3.86\pm0.43$
            & $-0.152\pm0.083$ \\
        & Test & $0.2589\pm0.0319$ & $3.90\pm0.50$
            & $-0.161\pm0.070$ \\
        \midrule
        ACS--IS & Train & $0.8068\pm0.0395$ & $1.24\pm0.06$
            & $0.145\pm0.075$ \\
        & Test & $0.8077\pm0.0397$ & $1.24\pm0.06$
            & $0.143\pm0.082$ \\
        \midrule
        ACS--ENLL & Train & $0.7813\pm0.0381$ & $1.28\pm0.06$
            & $0.169\pm0.087$ \\
        & Test & $0.7830\pm0.0362$ & $1.28\pm0.06$
            & $0.158\pm0.101$ \\
        \bottomrule
    \end{tabular*}
\end{table}

Both ACS objectives produce substantially more concentrated selected-affinity
distributions than Router CE: mean concentration increases from approximately
$0.26$ to $0.78$--$0.81$, while the reciprocal effective count decreases from
approximately $3.9$ to $1.2$--$1.3$. The association with observed token CE
also changes from weakly negative under Router CE to weakly positive under
both ACS objectives. These shifts are stable between train and test, but are
descriptive and do not establish that increased concentration causes the
corresponding accuracy differences.
The effective count describes within-token weight concentration: all $K$
selected experts still execute, and this statistic does not measure expert
load concentration across tokens.

\subsection{Coefficient Scaling Across Supervision Depth}
\label{app:acs_depth_coefficient_scaling}

Equation~\ref{eq:averaged_training_losses} averages the ACS objective over the
supervised MoE layers. Holding the global coefficient fixed while increasing
the number of supervised layers therefore reduces the effective contribution
associated with each layer. We test whether increasing the global coefficient
with supervision depth recovers the behavior observed under narrower
supervision.

Table~\ref{tab:acs_depth_coefficient_scaling} shows that restoring per-layer
supervision strength improves all four ARC-Challenge arms relative to the fixed
global coefficient. Across the other datasets, ten of twelve
scope--objective arms are unchanged or improved. These
results show that supervision depth and coefficient cannot be interpreted
independently. At the same time, the excessive-scaling control in
Table~\ref{tab:acs_excessive_scaling_control} shows that a further tenfold
increase sharply reduces accuracy, identifying an intermediate operating
range rather than a monotonic benefit from stronger supervision.

\input{tables/acs_depth_coefficient_scaling}

%% file: tables/tes_hyperparameter_sensitivity.tex
\begin{table}[htbp]
    \setlength{\belowcaptionskip}{\abovecaptionskip}
    \setlength{\abovecaptionskip}{0pt}
    \caption{TES hyperparameter sensitivity on Granite ARC-Challenge with
    started-log IS supervision. Entries are epoch-3 means over seeds 42--44.
    The learning-rate panel fixes $\gamma=1$; the attenuation panel fixes the
    error-head learning rate at $10^{-5}$. Bold marks the best observed value
    within each panel and metric.}
    \label{tab:tes_hyperparameter_sensitivity}
    \centering
    \small
    \setlength{\tabcolsep}{2.5pt}
    \renewcommand{\arraystretch}{1.12}
    \begin{tabular*}{\linewidth}{@{\extracolsep{\fill}}llrr@{}}
        \toprule
        Parameter & Value & \shortstack{Accuracy\\(\%)}
            & \shortstack{Choice\\NLL} \\
        \midrule
        Error-head learning rate
            & $10^{-3}$ & $66.07$ & $0.9811$ \\
        & $10^{-4}$ & $66.60$ & $0.9691$ \\
        & $10^{-5}$ & $\mathbf{66.87}$ & $\mathbf{0.9673}$ \\
        & $5\times10^{-6}$ & $66.67$ & $\mathbf{0.9673}$ \\
        \midrule
        Attenuation scale $\gamma$
            & $0.5$ & $65.87$ & $0.9729$ \\
        & $1$ & $\mathbf{66.87}$ & $0.9673$ \\
        & $2$ & $66.00$ & $\mathbf{0.9654}$ \\
        \bottomrule
    \end{tabular*}
\end{table}

%% file: tables/acs_coefficient_sensitivity.tex
\begin{table}[t]
    \caption{ACS coefficient sensitivity (accuracy, \%) under final-layer
    supervision. Bold marks the higher mean within each matched pair.}
    \label{tab:acs_coefficient_sensitivity}
    \centering
    \scriptsize
    \begin{tabular*}{\columnwidth}{@{\extracolsep{\fill}}lllccccc@{}}
        \toprule
        Model & Affinity & Objective & $\lambda_m$
            & ARC-C & \shortstack{OpenBook\\QA} & SciQ & MedMCQA \\
        \midrule
        Granite & Trainable & IS & $10^{-3}$
            & $70.60$ & $73.93$ & $\mathbf{89.53}$ & $40.67$ \\
        & & & $10^{-2}$
            & $\mathbf{71.20}$ & $\mathbf{74.53}$ & $88.80$ & $\mathbf{41.07}$ \\
        & & ENLL & $10^{-3}$
            & $69.87$ & $73.53$ & $\mathbf{89.20}$ & $\mathbf{42.13}$ \\
        & & & $10^{-2}$
            & $\mathbf{70.47}$ & $\mathbf{75.20}$ & $89.00$ & $41.47$ \\
        \midrule
        OLMoE & Frozen & IS & $10^{-3}$
            & $62.73$ & $\mathbf{69.93}$ & $88.87$ & $\mathbf{44.20}$ \\
        & & & $10^{-2}$
            & $\mathbf{63.40}$ & $69.80$ & $\mathbf{89.67}$ & $44.07$ \\
        & & ENLL & $10^{-3}$
            & $62.93$ & $\mathbf{70.20}$ & $88.73$ & $\mathbf{44.53}$ \\
        & & & $10^{-2}$
            & $\mathbf{63.93}$ & $69.20$ & $\mathbf{90.07}$ & $44.13$ \\
        \midrule
        OLMoE & Trainable & IS & $10^{-3}$
            & $63.60$ & $\mathbf{69.33}$ & $89.07$ & $\mathbf{44.53}$ \\
        & & & $10^{-2}$
            & $\mathbf{64.20}$ & $69.27$ & $\mathbf{89.47}$ & $43.73$ \\
        & & ENLL & $10^{-3}$
            & $63.80$ & $\mathbf{71.00}$ & $88.87$ & $\mathbf{44.27}$ \\
        & & & $10^{-2}$
            & $\mathbf{64.47}$ & $69.60$ & $\mathbf{89.80}$ & $43.67$ \\
        \bottomrule
    \end{tabular*}
\end{table}

\begin{table}[t]
    \caption{Complete frozen-affinity OLMoE ARC-Challenge coefficient screen.}
    \label{tab:olmoe_acs_coefficient_screen}
    \centering
    \scriptsize
    \begin{tabular*}{\columnwidth}{@{\extracolsep{\fill}}lcccc@{}}
        \toprule
        Objective & $10^{-4}$ & $10^{-3}$ & $10^{-2}$ & $10^{-1}$ \\
        \midrule
        ACS--IS
            & $63.27\pm0.12$ & $62.73\pm0.31$
            & $\mathbf{63.40\pm0.40}$ & $59.00\pm0.40$ \\
        ACS--ENLL
            & $62.33\pm0.46$ & $62.93\pm0.64$
            & $\mathbf{63.93\pm0.76}$ & $58.33\pm1.21$ \\
        \bottomrule
    \end{tabular*}
\end{table}

%% file: tables/affinity_router_layer_ablation.tex
\begin{table}[!ht]
    \setlength{\belowcaptionskip}{\abovecaptionskip}
    \setlength{\abovecaptionskip}{0pt}
    \caption{Answer-choice accuracy (\%; mean $\pm$ sample standard
    deviation over seeds 42, 43, and 45) under trainable native-affinity layer
    scopes. Granite values use the fixed epoch-3 checkpoint and
    $\lambda_m=10^{-3}$. OLMoE ACS values marked
    $^{\ast}$ use the checkpoint selected by minimum validation NLL; ACS rows
    use $\lambda_m=10^{-2}$. Bold
    marks the best evaluated scope within each model, method, and dataset.}
    \label{tab:affinity_router_layer_ablation}
    \centering
    \small
    \setlength{\tabcolsep}{2.5pt}
    \renewcommand{\arraystretch}{1.12}
    \resizebox{\columnwidth}{!}{%
    \begin{tabular}{@{}lllcccc@{}}
        \toprule
        Model & Method & \shortstack{ACS and affinity-\\training scope}
            & ARC-Challenge & OpenBookQA & SciQ & MedMCQA \\
        \midrule
        Granite 3.1 & Router CE
            & Final 1 & $\mathbf{67.53\pm1.14}$ & $\mathbf{74.00\pm0.72}$
            & $88.60\pm0.40$ & $\mathbf{41.73\pm1.22}$ \\
        & & First 8 & $65.60\pm3.02$ & $70.93\pm2.61$
            & $\mathbf{89.07\pm0.23}$ & $40.87\pm0.42$ \\
        & & Last 8 & $65.80\pm1.51$ & $71.07\pm0.61$
            & $88.07\pm0.61$ & $37.73\pm0.90$ \\
        & & All 32 & $61.60\pm4.20$ & $61.60\pm3.40$
            & $86.27\pm0.42$ & $30.80\pm3.34$ \\
        \cmidrule(lr){2-7}
        & ACS-ENLL
            & Final 1 & $\mathbf{69.87\pm1.33}$ & $\mathbf{73.53\pm0.90}$
            & $\mathbf{89.20\pm0.92}$ & $\mathbf{42.13\pm1.27}$ \\
        & & First 8 & $65.00\pm1.04$ & $71.33\pm1.10$
            & $88.87\pm1.53$ & $40.53\pm2.00$ \\
        & & Last 8 & $68.40\pm1.31$ & $69.60\pm3.67$
            & $88.40\pm0.80$ & $38.60\pm0.72$ \\
        & & All 32 & $63.60\pm0.92$ & $61.13\pm9.99$
            & $71.40\pm26.33$ & $35.87\pm1.10$ \\
        \cmidrule(lr){2-7}
        & ACS-IS
            & Final 1 & $\mathbf{70.60\pm1.80}$ & $\mathbf{73.93\pm0.31}$
            & $\mathbf{89.53\pm0.42}$ & $\mathbf{40.67\pm1.45}$ \\
        & & First 8 & $67.73\pm1.27$ & $70.00\pm1.40$
            & $88.00\pm0.69$ & $38.80\pm1.44$ \\
        & & Last 8 & $68.47\pm1.55$ & $70.20\pm0.35$
            & $88.67\pm1.03$ & $37.40\pm1.22$ \\
        & & All 32 & $60.13\pm10.34$ & $68.87\pm1.81$
            & $87.73\pm0.81$ & $35.00\pm2.31$ \\
        \midrule
        OLMoE & Router CE
            & Final 1 & $63.53\pm0.58^{\ast}$ & $\mathbf{71.87\pm0.42}^{\ast}$
            & $\mathbf{88.80\pm0.20}^{\ast}$ & $\mathbf{44.47\pm0.46}^{\ast}$ \\
        & & First 8 & $63.07\pm0.90^{\ast}$ & $71.20\pm0.20^{\ast}$
            & $88.73\pm0.23^{\ast}$ & $43.87\pm0.99^{\ast}$ \\
        & & Last 8 & $63.53\pm0.23^{\ast}$ & $70.40\pm0.53^{\ast}$
            & $\mathbf{88.80\pm0.20}^{\ast}$ & $43.47\pm0.64^{\ast}$ \\
        & & All 16 & $\mathbf{64.67\pm1.40}^{\ast}$ & $70.53\pm0.95^{\ast}$
            & $88.73\pm0.31^{\ast}$ & $43.60\pm0.87^{\ast}$ \\
        \cmidrule(lr){2-7}
        & ACS-ENLL
            & Final 1 & $\mathbf{64.47\pm0.70}^{\ast}$ & $69.60\pm0.53^{\ast}$
            & $\mathbf{89.80\pm0.20}^{\ast}$ & $43.67\pm0.70^{\ast}$ \\
        & & First 8 & $\mathbf{64.47\pm0.83}^{\ast}$ & $70.40\pm0.20^{\ast}$
            & $88.60\pm0.20^{\ast}$ & $43.87\pm1.03^{\ast}$ \\
        & & Last 8 & $64.13\pm0.31^{\ast}$ & $69.60\pm0.80^{\ast}$
            & $88.67\pm0.23^{\ast}$ & $\mathbf{44.20\pm0.40}^{\ast}$ \\
        & & All 16 & $63.47\pm0.12^{\ast}$ & $\mathbf{70.80\pm0.53}^{\ast}$
            & $88.07\pm0.23^{\ast}$ & $43.13\pm1.10^{\ast}$ \\
        \cmidrule(lr){2-7}
        & ACS-IS
            & Final 1 & $64.20\pm0.20^{\ast}$ & $69.27\pm0.42^{\ast}$
            & $\mathbf{89.47\pm0.42}^{\ast}$ & $43.73\pm0.58^{\ast}$ \\
        & & First 8 & $64.07\pm1.29^{\ast}$ & $69.80\pm0.20^{\ast}$
            & $89.33\pm0.50^{\ast}$ & $\mathbf{44.27\pm0.70}^{\ast}$ \\
        & & Last 8 & $63.73\pm0.23^{\ast}$ & $\mathbf{70.00\pm0.87}^{\ast}$
            & $88.93\pm0.46^{\ast}$ & $44.07\pm0.50^{\ast}$ \\
        & & All 16 & $\mathbf{64.27\pm0.90}^{\ast}$ & $\mathbf{70.00\pm0.20}^{\ast}$
            & $88.60\pm0.35^{\ast}$ & $43.40\pm0.80^{\ast}$ \\
        \bottomrule
    \end{tabular}
    }
\end{table}

%% file: tables/acs_depth_coefficient_scaling.tex
\begin{table}[t]
    \caption{ARC-Challenge excessive-scaling control (accuracy, \%).}
    \label{tab:acs_excessive_scaling_control}
    \centering
    \scriptsize
    \renewcommand{\arraystretch}{1.12}
    \begin{tabular*}{\columnwidth}{@{\extracolsep{\fill}}llrrr@{}}
        \toprule
        Scope & Objective & \shortstack{Depth-\\scaled}
            & \shortstack{$10\times$\\scaled} & Change \\
        \midrule
        Last 8 & IS & $69.27\pm0.64$ & $57.73\pm2.32$ & $-11.53$ \\
        & ENLL & $68.80\pm1.06$ & $56.13\pm1.22$ & $-12.67$ \\
        All 32 & IS & $67.20\pm3.47$ & $48.60\pm0.92$ & $-18.60$ \\
        & ENLL & $69.00\pm1.11$ & $48.13\pm0.50$ & $-20.87$ \\
        \bottomrule
    \end{tabular*}
\end{table}

\begin{table}[t]
    \caption{Granite epoch-3 accuracy (\%) under fixed and depth-scaled ACS
    coefficients. Bold marks the higher matched mean.}
    \label{tab:acs_depth_coefficient_scaling}
    \scriptsize
    \begin{tabular*}{\columnwidth}{@{\extracolsep{\fill}}lllrrr@{}}
        \toprule
        Dataset & Scope & Objective & \shortstack{Fixed\\$10^{-3}$}
            & \shortstack{Depth-\\scaled} & Change \\
        \midrule
        ARC-C & Last 8 & IS & $68.47\pm1.55$ & $\mathbf{69.27\pm0.64}$ & $+0.80$ \\
        & & ENLL & $68.40\pm1.31$ & $\mathbf{68.80\pm1.06}$ & $+0.40$ \\
        & All 32 & IS & $60.13\pm10.34$ & $\mathbf{67.20\pm3.47}$ & $+7.07$ \\
        & & ENLL & $63.60\pm0.92$ & $\mathbf{69.00\pm1.11}$ & $+5.40$ \\
        \midrule
        OpenBookQA & Last 8 & IS & $70.20\pm0.35$ & $\mathbf{70.27\pm0.12}$ & $+0.07$ \\
        & & ENLL & $69.60\pm3.67$ & $\mathbf{70.20\pm0.53}$ & $+0.60$ \\
        & All 32 & IS & $\mathbf{68.87\pm1.81}$ & $59.20\pm9.53$ & $-9.67$ \\
        & & ENLL & $61.13\pm9.99$ & $\mathbf{65.00\pm1.22}$ & $+3.87$ \\
        \midrule
        SciQ & Last 8 & IS & $88.67\pm1.03$ & $\mathbf{89.27\pm0.50}$ & $+0.60$ \\
        & & ENLL & $88.40\pm0.80$ & $\mathbf{88.87\pm0.42}$ & $+0.47$ \\
        & All 32 & IS & $\mathbf{87.73\pm0.81}$ & $87.53\pm1.01$ & $-0.20$ \\
        & & ENLL & $71.40\pm26.33$ & $\mathbf{88.00\pm0.92}$ & $+16.60$ \\
        \midrule
        MedMCQA & Last 8 & IS & $37.40\pm1.22$ & $\mathbf{38.00\pm4.78}$ & $+0.60$ \\
        & & ENLL & $38.60\pm0.72$ & $\mathbf{38.80\pm1.22}$ & $+0.20$ \\
        & All 32 & IS & $35.00\pm2.31$ & $\mathbf{37.27\pm0.81}$ & $+2.27$ \\
        & & ENLL & $35.87\pm1.10$ & $\mathbf{37.40\pm2.43}$ & $+1.53$ \\
        \bottomrule
    \end{tabular*}
\end{table}

%% file: smogu.bib
@misc{ibm2024granite31,
  title        = {{Granite-3.1-3B-A800M-Base}},
  author       = {{IBM Granite Team}},
  year         = {2024},
  howpublished = {Hugging Face model card},
  url          = {https://huggingface.co/ibm-granite/granite-3.1-3b-a800m-base}
}

@article{muennighoff2024olmoe,
  title         = {{OLMoE}: Open Mixture-of-Experts Language Models},
  author        = {Niklas Muennighoff and Luca Soldaini and Dirk Groeneveld and Kyle Lo and Jacob Morrison and Sewon Min and Weijia Shi and Pete Walsh and Oyvind Tafjord and Nathan Lambert and Yuling Gu and Shane Arora and Akshita Bhagia and Dustin Schwenk and David Wadden and Alexander Wettig and Binyuan Hui and Tim Dettmers and Douwe Kiela and Ali Farhadi and Noah A. Smith and Pang Wei Koh and Amanpreet Singh and Hannaneh Hajishirzi},
  journal       = {arXiv preprint arXiv:2409.02060},
  year          = {2024},
  eprint        = {2409.02060},
  archiveprefix = {arXiv},
  primaryclass  = {cs.CL},
  url           = {https://arxiv.org/abs/2409.02060}
}

@article{clark2018arc,
  title         = {Think You Have Solved Question Answering? Try {ARC}, the {AI2} Reasoning Challenge},
  author        = {Peter Clark and Isaac Cowhey and Oren Etzioni and Tushar Khot and Ashish Sabharwal and Carissa Schoenick and Oyvind Tafjord},
  journal       = {arXiv preprint arXiv:1803.05457},
  year          = {2018},
  eprint        = {1803.05457},
  archiveprefix = {arXiv},
  primaryclass  = {cs.AI},
  url           = {https://arxiv.org/abs/1803.05457}
}

@article{mihaylov2018openbookqa,
  title         = {Can a Suit of Armor Conduct Electricity? A New Dataset for Open Book Question Answering},
  author        = {Todor Mihaylov and Peter Clark and Tushar Khot and Ashish Sabharwal},
  journal       = {arXiv preprint arXiv:1809.02789},
  year          = {2018},
  eprint        = {1809.02789},
  archiveprefix = {arXiv},
  primaryclass  = {cs.CL},
  url           = {https://arxiv.org/abs/1809.02789}
}

@inproceedings{welbl2017sciq,
  title     = {Crowdsourcing Multiple Choice Science Questions},
  author    = {Welbl, Johannes and Liu, Nelson F. and Gardner, Matt},
  booktitle = {Proceedings of the 3rd Workshop on Noisy User-generated Text},
  year      = {2017},
  pages     = {94--106},
  publisher = {Association for Computational Linguistics},
  doi       = {10.18653/v1/w17-4413},
  url       = {https://aclanthology.org/W17-4413/}
}

@article{pal2022medmcqa,
  title         = {{MedMCQA}: A Large-scale Multi-Subject Multi-Choice Dataset for Medical Domain Question Answering},
  author        = {Ankit Pal and Logesh Kumar Umapathi and Malaikannan Sankarasubbu},
  journal       = {arXiv preprint arXiv:2203.14371},
  year          = {2022},
  eprint        = {2203.14371},
  archiveprefix = {arXiv},
  primaryclass  = {cs.CL},
  url           = {https://arxiv.org/abs/2203.14371}
}

@inproceedings{lepikhin2021gshard,
  title         = {{GShard}: Scaling Giant Models with Conditional Computation and Automatic Sharding},
  author        = {Dmitry Lepikhin and HyoukJoong Lee and Yuanzhong Xu and Dehao Chen and Orhan Firat and Yanping Huang and Maxim Krikun and Noam Shazeer and Zhifeng Chen},
  booktitle     = {International Conference on Learning Representations},
  year          = {2021},
  eprint        = {2006.16668},
  archiveprefix = {arXiv},
  primaryclass  = {cs.CL},
  url           = {https://arxiv.org/abs/2006.16668}
}

@article{fedus2022switch,
  title   = {Switch Transformers: Scaling to Trillion Parameter Models with Simple and Efficient Sparsity},
  author  = {William Fedus and Barret Zoph and Noam Shazeer},
  journal = {Journal of Machine Learning Research},
  year    = {2022},
  volume  = {23},
  number  = {120},
  pages   = {1--39},
  url     = {https://www.jmlr.org/papers/v23/21-0998.html}
}

@article{jiang2024mixtral,
  title         = {Mixtral of Experts},
  author        = {Albert Q. Jiang and Alexandre Sablayrolles and Antoine Roux and Arthur Mensch and Blanche Savary and Chris Bamford and Devendra Singh Chaplot and Diego de las Casas and Emma Bou Hanna and Florian Bressand and Gianna Lengyel and Guillaume Bour and Guillaume Lample and Lélio Renard Lavaud and Lucile Saulnier and Marie-Anne Lachaux and Pierre Stock and Sandeep Subramanian and Sophia Yang and Szymon Antoniak and Teven Le Scao and Théophile Gervet and Thibaut Lavril and Thomas Wang and Timothée Lacroix and William El Sayed},
  journal       = {arXiv preprint arXiv:2401.04088},
  year          = {2024},
  eprint        = {2401.04088},
  archiveprefix = {arXiv},
  primaryclass  = {cs.LG},
  url           = {https://arxiv.org/abs/2401.04088}
}

@article{dai2024deepseekmoe,
  title         = {{DeepSeekMoE}: Towards Ultimate Expert Specialization in Mixture-of-Experts Language Models},
  author        = {Damai Dai and Chengqi Deng and Chenggang Zhao and R. X. Xu and Huazuo Gao and Deli Chen and Jiashi Li and Wangding Zeng and Xingkai Yu and Y. Wu and Zhenda Xie and Y. K. Li and Panpan Huang and Fuli Luo and Chong Ruan and Zhifang Sui and Wenfeng Liang},
  journal       = {arXiv preprint arXiv:2401.06066},
  year          = {2024},
  eprint        = {2401.06066},
  archiveprefix = {arXiv},
  primaryclass  = {cs.CL},
  url           = {https://arxiv.org/abs/2401.06066}
}

@article{aviv2025mogu,
  title         = {{MoGU}: Mixture-of-Gaussians with Uncertainty-based Gating for Time Series Forecasting},
  author        = {Gilad Aviv and Jacob Goldberger and Yoli Shavit},
  journal       = {arXiv preprint arXiv:2510.07459},
  year          = {2025},
  eprint        = {2510.07459},
  archiveprefix = {arXiv},
  primaryclass  = {cs.LG},
  url           = {https://arxiv.org/abs/2510.07459}
}

@inproceedings{li2026vmoer,
  title         = {Variational Routing: A Scalable Bayesian Framework for Calibrated Mixture-of-Experts Transformers},
  author        = {Albus Yizhuo Li and Matthew Wicker},
  booktitle     = {Proceedings of the 43rd International Conference on Machine Learning},
  year          = {2026},
  eprint        = {2603.09453},
  archiveprefix = {arXiv},
  primaryclass  = {cs.LG},
  url           = {https://arxiv.org/abs/2603.09453}
}

@inproceedings{chen2026uar,
  title     = {Uncertainty-Aware Routing for Principled Alignment with {MoE} Dynamics},
  author    = {Chen, Yilong and Shang, Junyuan and Feng, Yuchen and Zhang, Zhenyu and Gu, Naibin and Wang, Ziqi and Liu, Tingwen and Wang, Shuohuan and Sun, Yu and Wu, Hua and Wang, Haifeng},
  editor    = {Liakata, Maria and Moreira, Viviane P. and Zhang, Jiajun and Jurgens, David},
  booktitle = {Proceedings of the 64th Annual Meeting of the Association for Computational Linguistics (Volume 1: Long Papers)},
  year      = {2026},
  month     = {jul},
  pages     = {38865--38880},
  publisher = {Association for Computational Linguistics},
  address   = {San Diego, California, United States},
  doi       = {10.18653/v1/2026.acl-long.1801},
  url       = {https://aclanthology.org/2026.acl-long.1801/}
}

@article{shihab2026grmoe,
  title         = {Grassmannian Mixture-of-Experts: Concentration-Controlled Routing on Subspace Manifolds},
  author        = {Ibne Farabi Shihab and Sanjeda Akter and Anuj Sharma},
  journal       = {arXiv preprint arXiv:2602.17798},
  year          = {2026},
  eprint        = {2602.17798},
  archiveprefix = {arXiv},
  primaryclass  = {cs.LG},
  url           = {https://arxiv.org/abs/2602.17798}
}

@article{saliencro2026vimole,
  title         = {Uncertainty Is Not Enough: Value-of-Information Routing for Mixtures of {LoRA} Experts},
  author        = {Tom Saliencro and Rohan Desai and Priya Nair and Maya Lindqvist and Daniel Whitmore},
  journal       = {arXiv preprint arXiv:2608.02528},
  year          = {2026},
  eprint        = {2608.02528},
  archiveprefix = {arXiv},
  primaryclass  = {cs.LG},
  url           = {https://arxiv.org/abs/2608.02528}
}

@article{shazeer2017outrageously,
  title         = {Outrageously large neural networks: The sparsely-gated mixture-of-experts layer},
  author        = {Noam Shazeer and Azalia Mirhoseini and Krzysztof Maziarz and Andy Davis and Quoc Le and Geoffrey Hinton and Jeff Dean},
  journal       = {arXiv preprint arXiv:1701.06538},
  year          = {2017},
  eprint        = {1701.06538},
  archiveprefix = {arXiv},
  primaryclass  = {cs.LG},
  url           = {https://arxiv.org/abs/1701.06538}
}

@inproceedings{zhou2022expertchoice,
  title     = {Mixture-of-Experts with Expert Choice Routing},
  author    = {Zhou, Yanqi and Lei, Tao and Liu, Hanxiao and Du, Nan and Huang, Yanping and Zhao, Vincent and Dai, Andrew and Chen, Zhifeng and Le, Quoc V and Laudon, James},
  editor    = {S. Koyejo and S. Mohamed and A. Agarwal and D. Belgrave and K. Cho and A. Oh},
  booktitle = {Advances in Neural Information Processing Systems 35},
  year      = {2022},
  volume    = {35},
  pages     = {7103--7114},
  publisher = {Neural Information Processing Systems Foundation, Inc. (NeurIPS)},
  doi       = {10.52202/068431-0515},
  url       = {https://proceedings.neurips.cc/paper_files/paper/2022/hash/2f00ecd787b432c1d36f3de9800728eb-Abstract-Conference.html}
}

@inproceedings{wang2025remoe,
  title         = {{ReMoE}: Fully Differentiable Mixture-of-Experts with {ReLU} Routing},
  author        = {Ziteng Wang and Jun Zhu and Jianfei Chen},
  booktitle     = {International Conference on Learning Representations},
  year          = {2025},
  eprint        = {2412.14711},
  archiveprefix = {arXiv},
  primaryclass  = {cs.LG},
  url           = {https://arxiv.org/abs/2412.14711}
}

@inproceedings{lv2026erc,
  title     = {Coupling Experts and Routers in Mixture-of-Experts via an Auxiliary Loss},
  author    = {Lv, Ang and Ma, Jin and Ma, Yiyuan and Qiao, Siyuan},
  editor    = {C. Vondrick and B. Hariharan and C. Raffel and L. Pinto and D. Yang and A. Faust},
  booktitle = {International Conference on Learning Representations},
  year      = {2026},
  volume    = {2026},
  pages     = {75251--75271},
  url       = {https://proceedings.iclr.cc/paper_files/paper/2026/hash/79ff18412c1a5816f071a796375abc3d-Abstract-Conference.html}
}

@inproceedings{li2026expertdivergence,
  title         = {Expert Divergence Learning for {MoE}-based Language Models},
  author        = {Jiaang Li and Haibin Chen and Langming Liu and Yujin Yuan and Yadao Wang and Yizhen Zhang and Chengting Yu and Xin Tong and Weidong Zhang and Shilei Liu and Wenbo Su and Bo Zheng},
  booktitle     = {International Conference on Learning Representations},
  year          = {2026},
  eprint        = {2603.00054},
  archiveprefix = {arXiv},
  primaryclass  = {cs.LG},
  url           = {https://arxiv.org/abs/2603.00054}
}

@article{yoon2026misrouted,
  title         = {When Are Experts Misrouted? Counterfactual Routing Analysis in Mixture-of-Experts Language Models},
  author        = {Youngsik Yoon and Siwei Wang and Wei Chen and Jungseul Ok},
  journal       = {arXiv preprint arXiv:2605.07260},
  year          = {2026},
  eprint        = {2605.07260},
  archiveprefix = {arXiv},
  primaryclass  = {cs.LG},
  url           = {https://arxiv.org/abs/2605.07260}
}

@inproceedings{yoo2019learning,
  title     = {Learning Loss for Active Learning},
  author    = {Yoo, Donggeun and Kweon, In So},
  booktitle = {Proceedings of the IEEE/CVF Conference on Computer Vision and Pattern Recognition (CVPR)},
  year      = {2019},
  month     = {June},
  pages     = {93--102},
  url       = {https://openaccess.thecvf.com/content_CVPR_2019/html/Yoo_Learning_Loss_for_Active_Learning_CVPR_2019_paper.html}
}

@inproceedings{ding2024hybrid,
  title         = {Hybrid {LLM}: Cost-Efficient and Quality-Aware Query Routing},
  author        = {Dujian Ding and Ankur Mallick and Chi Wang and Robert Sim and Subhabrata Mukherjee and Victor Ruhle and Laks V. S. Lakshmanan and Ahmed Hassan Awadallah},
  booktitle     = {International Conference on Learning Representations},
  year          = {2024},
  eprint        = {2404.14618},
  archiveprefix = {arXiv},
  primaryclass  = {cs.LG},
  url           = {https://arxiv.org/abs/2404.14618}
}

@inproceedings{huang2024harder,
  title     = {Harder Task Needs More Experts: Dynamic Routing in {MoE} Models},
  author    = {Huang, Quzhe and An, Zhenwei and Zhuang, Nan and Tao, Mingxu and Zhang, Chen and Jin, Yang and Xu, Kun and Xu, Kun and Chen, Liwei and Huang, Songfang and Feng, Yansong},
  booktitle = {Proceedings of the 62nd Annual Meeting of the Association for Computational Linguistics (Volume 1: Long Papers)},
  year      = {2024},
  pages     = {12883--12895},
  publisher = {Association for Computational Linguistics},
  doi       = {10.18653/v1/2024.acl-long.696},
  url       = {https://aclanthology.org/2024.acl-long.696/}
}

@article{nishu2025dynamoe,
  title         = {From Dense to Dynamic: Token-Difficulty Driven {MoE}fication of Pre-Trained {LLM}s},
  author        = {Kumari Nishu and Sachin Mehta and Samira Abnar and Mehrdad Farajtabar and Maxwell Horton and Mahyar Najibi and Moin Nabi and Minsik Cho and Devang Naik},
  journal       = {arXiv preprint arXiv:2502.12325},
  year          = {2025},
  eprint        = {2502.12325},
  archiveprefix = {arXiv},
  primaryclass  = {cs.CL},
  url           = {https://arxiv.org/abs/2502.12325}
}

@inproceedings{zhao2025adak,
  title     = {{Ada-K} Routing: Boosting the Efficiency of {MoE}-based {LLM}s},
  author    = {Zhao, Zijia and Guo, Longteng and Cheng, Jie and Gao, Xuange and Huang, Hua and Liu, Jing},
  editor    = {Y. Yue and A. Garg and N. Peng and F. Sha and R. Yu},
  booktitle = {International Conference on Learning Representations},
  year      = {2025},
  volume    = {2025},
  pages     = {89619--89635},
  url       = {https://proceedings.iclr.cc/paper_files/paper/2025/hash/df22a19686a558e74f038e6277a51f68-Abstract-Conference.html}
}

@inproceedings{itakura1968analysis,
  title     = {Analysis Synthesis Telephony Based on the Maximum Likelihood Method},
  author    = {Itakura, Fumitada and Saito, Shuzo},
  booktitle = {Proceedings of the 6th International Congress on Acoustics},
  year      = {1968},
  pages     = {C17--C20},
  url       = {https://cir.nii.ac.jp/crid/1570854175842518528}
}

@article{fevotte2009nonnegative,
  title     = {Nonnegative Matrix Factorization with the {Itakura--Saito} Divergence: With Application to Music Analysis},
  author    = {Févotte, Cédric and Bertin, Nancy and Durrieu, Jean-Louis},
  journal   = {Neural Computation},
  year      = {2009},
  month     = {Mar},
  volume    = {21},
  number    = {3},
  pages     = {793--830},
  publisher = {MIT Press},
  doi       = {10.1162/neco.2008.04-08-771},
  url       = {https://doi.org/10.1162/neco.2008.04-08-771}
}

@book{casella2002statistical,
  title     = {Statistical Inference},
  author    = {Casella, George and Berger, Roger L.},
  year      = {2002},
  publisher = {Duxbury},
  address   = {Belmont, CA},
  url       = {https://ci.nii.ac.jp/ncid/BA5282379X},
  edition   = {2},
  isbn      = {9780534243128}
}

@article{rocke2003approximate,
  title     = {Approximate Variance-Stabilizing Transformations for Gene-Expression Microarray Data},
  author    = {Rocke, David M. and Durbin, Blythe},
  journal   = {Bioinformatics},
  year      = {2003},
  month     = {May},
  volume    = {19},
  number    = {8},
  pages     = {966--972},
  publisher = {Oxford University Press (OUP)},
  doi       = {10.1093/bioinformatics/btg107},
  url       = {https://doi.org/10.1093/bioinformatics/btg107}
}

@inproceedings{renyi1961measures,
  title     = {On Measures of Entropy and Information},
  author    = {R{\'e}nyi, Alfr{\'e}d},
  booktitle = {Proceedings of the Fourth Berkeley Symposium on Mathematical Statistics and Probability},
  year      = {1961},
  volume    = {1},
  pages     = {547--561},
  publisher = {University of California Press},
  url       = {https://cir.nii.ac.jp/crid/1571417126122876416}
}

@article{hill1973diversity,
  title     = {Diversity and Evenness: A Unifying Notation and Its Consequences},
  author    = {Hill, M. O.},
  journal   = {Ecology},
  year      = {1973},
  month     = {Mar},
  volume    = {54},
  number    = {2},
  pages     = {427--432},
  publisher = {Wiley},
  doi       = {10.2307/1934352},
  url       = {https://doi.org/10.2307/1934352}
}
